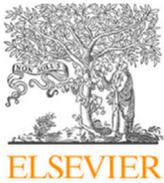

Contents lists available at ScienceDirect

# Information Fusion

journal homepage: www.elsevier.com/locate/inffus

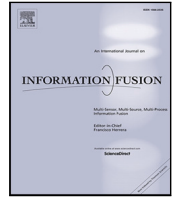

Full length article

# Advancements in point cloud-based 3D defect classification and segmentation for industrial systems: A comprehensive survey

Anju Rani *, Daniel Ortiz-Arroyo, Petar Durdevic

*Department of Energy, Aalborg University, Niels Bohrs Vej 8, Esbjerg, 6700, Denmark*

A R T I C L E   I N F O

A B S T R A C T

In recent years, 3D point clouds (PCs) have gained significant attention due to their diverse applications across various fields, such as computer vision (CV), condition monitoring (CM), virtual reality, robotics, autonomous driving, etc. Deep learning (DL) has proven effective in leveraging 3D PCs to address various challenges encountered in 2D vision. However, applying deep neural networks (DNNs) to process 3D PCs presents unique challenges. This paper provides an in-depth review of recent advancements in DL-based industrial CM using 3D PCs, with a specific focus on defect shape classification and segmentation within industrial applications. Recognizing the crucial role of these aspects in industrial maintenance, the paper offers insightful observations on the strengths and limitations of the reviewed DL-based PC processing methods. This knowledge synthesis aims to contribute to understanding and enhancing CM processes, particularly within the framework of remaining useful life (RUL), in industrial systems.

## 1. Introduction

Condition monitoring (CM) is vital in ensuring the longevity and proper maintenance of structures, such as bridges, buildings, industrial facilities, and infrastructure. Traditional visual inspection has been the predominant approach for CM applications. However, two-dimensional images face limitations in providing depth information and relative object positions, which is crucial for tasks involving spatial details, such as autonomous driving, virtual reality, and robotics. The emergence of 3D acquisition technologies, including depth sensors and 3D scanners, has effectively addressed this limitation by facilitating the extraction of detailed 3D information. The utilization of 3D data offers a significantly improved understanding of objects compared to traditional 2D images, making them a valuable tool for industrial CM applications. In recent years, there has been a growing emphasis among researchers on harnessing 3D scanned objects for defect detection and segmentation in industrial applications [1–4]. The representation of 3D data can be in various forms, including depth images, PCs, meshes, and volumetric grids. PC representation stands out for preserving the original geometric features in 3D space without any discretization, making it the preferred choice in many applications. The PC consists of unstructured 3D vectors, where each point represents a vector indicating its 3D coordinates (XYZ) with additional feature channels such as color (RGB values), intensity, and surface normals. Also, the PC exhibits properties like unstructured points, interaction among points, and invariance under transformation. These characteristics contribute to the flexibility and adaptability of PC representation in capturing complex geometric structures, which is crucial for identifying and characterizing defects in industrial systems.

In the last decade, DL has emerged as the most influential technique in the field of 2D-CV such as image recognition, object detection, and segmentation. However, the application of DL to 3D PC data presents unique challenges due to the unstructured, high-dimensional, and disordered nature of PCs. Traditional convolutional networks designed for regular grids may not be directly applicable to PCs. Therefore, raw PC data is pre-processed to make it compatible with DL algorithms. This involves steps such as noise removal, data cleaning, down-sampling, and normalization to enhance and ensure data consistency. Later, various network architectures, including convolutional neural networks (CNNs) [5–7], graph neural networks (GNNs) [8–10], or hybrid networks [11–13], can be used for specific tasks such as 3D classification and segmentation. The DL model is then trained using annotated PC data. This involves feeding the PC into the network, computing the loss between ground truth labels and predicted labels, and then updating model parameters through back-propagation. The training stage often requires a large input dataset, prompting the use of data augmentation techniques to improve generalization. After training and evaluation, the model can be used for inference on new, unseen PC data, followed by post-processing steps to refine the model output. A taxonomy of existing DL methods for processing 3D PCs is shown in Fig. 1.

* Corresponding author.
*E-mail addresses:* aran@et.aau.dk (A. Rani), doa@et.aau.dk (D. Ortiz-Arroyo), pdl@et.aau.dk (P. Durdevic).

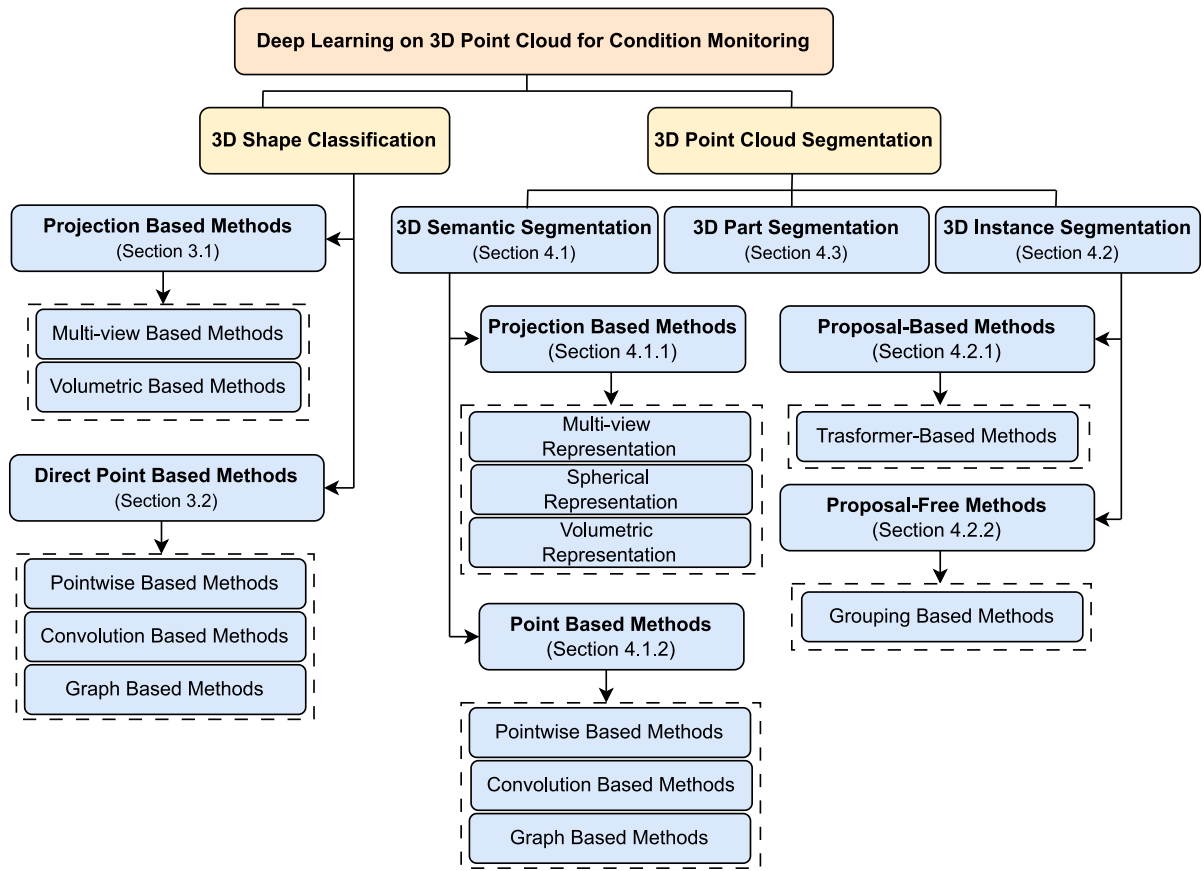

**Fig. 1.** A taxonomy of DL methods for processing 3D PC data.

The paper provides a comprehensive review of DL methods applied to 3D PC data, with a specific emphasis on their applications in industrial CM settings. While previous reviews have explored DL techniques using standard datasets, this paper goes beyond by dissecting fundamental methodologies and recent advancements in 3D shape classification and segmentation, specifically catering to CM requirements in industrial CM applications. The reviewed methods are explicitly applicable to addressing the challenges encountered in industrial CM tasks, such as identifying and locating defects in equipment, infrastructure, and manufacturing processes. The review covers traditional and innovative approaches, shedding light on the inherent challenges and potential solutions in processing 3D PC data for CM applications in industrial settings. Additionally, it provides a detailed summary of existing DL methodologies for feature learning in 3D PCs, outlining their respective strengths and weaknesses. Including publicly available datasets relevant to 3D shape classification and object segmentation enhances the practical value of the discussion. Overall, the synthesis of existing knowledge in this review aims to identify gaps in the current understanding and pave the way for further innovations in the dynamic field of 3D PC data processing, offering valuable insights for researchers and practitioners working on advancing industrial CM capabilities. The key contributions of this review paper encompass the following aspects:

1. The paper thoroughly surveys the most recent advancements in DL-based 3D PCs applied to both traditional and CM applications. The discussion is categorized into two main domains—shape classification and 3D object segmentation.

2. The review systematically compares and summarizes recent methods for CM, with a specific focus on damage detection in industrial applications. This comparative analysis not only highlights the diverse approaches but also provides an insightful assessment of the strengths and limitations of each method, offering valuable guidance for researchers and practitioners.

3. The paper goes beyond the current state of the field by offering valuable insights into potential future research directions and applications in the realm of DL-based CM using 3D PCs. This forward-looking perspective aims to inspire and guide future research endeavors in the dynamic and evolving field.

The structure of this review paper is organized as follows: Section 2 discusses the existing datasets and evaluation metrics utilized for 3D PC classification and segmentation tasks. Section 3 focuses on DL methods used for 3D shape classification, unraveling the evolution and applications of these methodologies. In Section 4, an extensive survey is conducted on existing methods for 3D PC segmentation, including semantic segmentation, instance segmentation, and part segmentation. The review concludes in Section 5, synthesizing insights and outlining future research directions.

## 2. Background

### 2.1. 3D datasets

The availability of publicly accessible datasets plays a pivotal role in facilitating the analysis and comparison of various models in the domain of 3D PC applications. Researchers have curated diverse datasets specifically designed for tasks such as 3D shape classification, 3D object detection, and 3D PC segmentation. Table 1 provides a concise summary of these benchmark datasets and their descriptions. These datasets can be broadly categorized into two main types: real-world and synthetic datasets. In real-world datasets [18,29], the objects are occluded at varying levels, while some objects may contain background noise. On the other hand, objects in synthetic datasets [26,27] are without any occlusion and background noise, offering a controlled environment for experimentation.

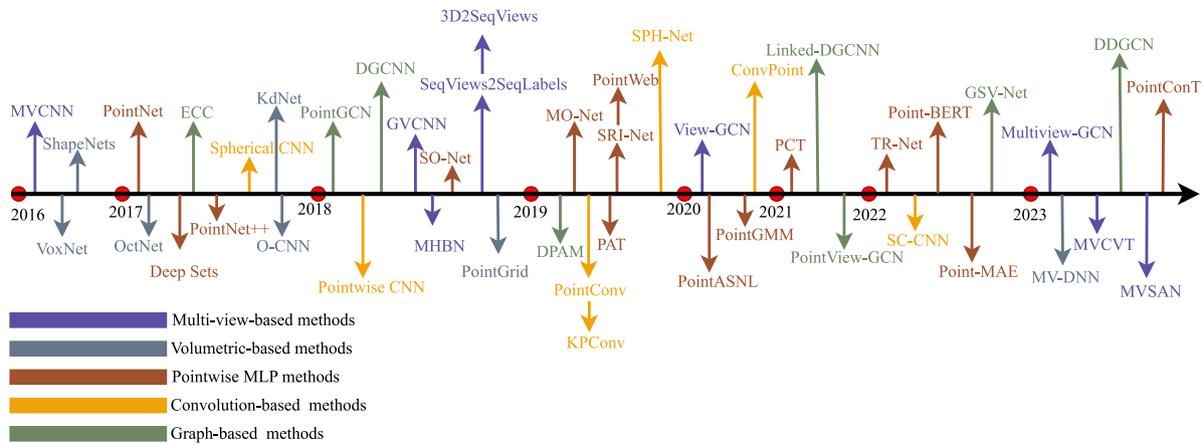

**Fig. 2.** Chronological overview of the most relevant DL-based 3D shape classification methods.

**Table 1**
Available benchmark PC dataset for classification and segmentation.

| Ref. | Dataset | Description | Year | Classes | Object/Point count | Classification | Segmentation |
|---|---|---|---|---|---|---|---|
| [14] | Oakland | Urban environment | 2009 | 44 | 1.6 M | ✓ | |
| [15] | ISPRS | Buildings, trees, and 3D building reconstruction | 2012 | 9 | 1.2 M | | ✓ |
| [16] | Paris-rue-Madame | Street in Paris | 2014 | 17 | 20 M | ✓ | ✓ |
| [17] | IQmulus | Dense urban environments | 2015 | 22 | 300 M | ✓ | ✓ |
| [18] | ScanNet | Indoor scenes | 2017 | 20 | 2.5 M | ✓ | ✓ |
| [19] | S3DIS | Structural elements | 2017 | 13 | 273 M | | ✓ |
| [20] | Semantic3D | Robotics, augmented reality and urban planning | 2017 | 9 | 4000 M | ✓ | |
| [21] | Paris-Lille-3D | Objects in urban environment | 2018 | 50 | 143 M | ✓ | |
| [22] | SematicKITTI | Autonomous driving | 2019 | 28 | 4549 M | | ✓ |
| [23] | Toronto | Urban roadways | 2020 | 9 | 78.3 M | | ✓ |
| [24] | DALES | Aerial geographical scan | 2020 | 9 | 505 M | ✓ | ✓ |
| [25] | nuScenes | Autonomous driving | 2020 | 7 | 5 B | | ✓ |
| [26] | ModelNet | CAD-generated objects | 2015 | 662 | 1.3 M | ✓ | |
| [27] | ShapeNet | CAD-generated objects | 2015 | 3135 | 300 M | ✓ | |
| [28] | ModelNet40-C | Corruption robustness | 2022 | 40 | 1.85 M | ✓ | |
| [29] | ScanObjectNN | Scanned indoor scenes | 2019 | 15 | 15,000 | ✓ | |
| [30] | STPLS3D | Synthetic and real aerial photogrammetry | 2022 | 20 | 15,888 | | ✓ |
| [31] | SUN RGB-D | 3D room layout and scenes | 2015 | 700 | 10,335 | | ✓ |
| [32] | Hypersim | Synthetic indoor images | 2021 | 461 | 77,400 | | ✓ |
| [33] | MVTec 3D-AD | 3D random objects | 2022 | 10 | 4000 | | ✓ |
| [34] | Real3D-AD | 3D random objects | 2023 | 12 | 1.3 M | | ✓ |

## 2.2. Evaluation metrics

Different evaluation metrics are employed in the literature to assess the performance of DL-based 3D PC processing tasks. For 3D shape classification, the most common performance criteria include *overall accuracy (OA)* and *mean class accuracy (mAcc)*, respectively. OA represents the mean accuracy for all test instances, while *mAcc* is the mean accuracy for all shape classes. In the case of 3D PC segmentation, *OA*, *mAcc*, *mean intersection over union (mIoU)* and *mean average precision (mAP)* are the most frequently used performance criteria. *OA* in this case represents the mean accuracy for PC segmentation, *mAcc* depicts the mean accuracy for different classes in segmentation, and *mIoU* Measures the overlap between predicted and ground truth segments. Particularly, *mAP* is used in instance segmentation of 3D PCs. These metrics provide a quantitative assessment of the performance of DL models across various 3D PC processing tasks. However, the appropriate metric is chosen based on the specific task and the desired aspects of performance to be evaluated.

## 3. Deep learning for 3D shape classification

The existing 3D shape classification methods can be broadly categorized into two major groups: projection-based methods and direct point-based methods. Fig. 2 depicts various milestone methods within these categories, showcasing the diversity of approaches discussed in the literature.

## 3.1. Projection-based methods

These methods typically involve projecting a 3D PC into 2D images, facilitating the application of well-established 2D image processing techniques for classification tasks. This category encompasses techniques leveraging multi-view or volumetric images to represent and analyze 3D shapes.

### 3.1.1. Multi-view based methods

This method captures 3D shape projections from multiple viewpoints and extracts features independently from each view. Traditional methods, such as CNNs, can be applied to each view to extract distinctive features, which are subsequently fused to classify the shape accurately. However, the effectiveness of these methods largely depends on the number of views selected for the classification. Multi-view CNN (MVCNN) [35] captures 3D shapes from various viewpoints, processing each view independently through CNNs (CNN1) to extract features. These features are then aggregated using a view-pooling layer, which collects information from all viewpoints. The pooled information is passed through three fully connected CNNs (CNN2) to produce a compact 3D shape descriptor. MVCNN's primary contribution is the synthesis of information from multiple viewpoints, which is particularly useful for industrial CM applications where the ability to accurately classify and locate defects in complex industrial systems is paramount. Here, the most important regions or saliency maps for each 3D shape $S$ are computed by back-propagation gradients of the class score. If $I_1, I_2, \ldots, I_n$ are the set of $n$ 2D views, then the viewpoints in 2D views are ranked based on their output score $F_g$ for its ground truth class $g$. The saliency maps are defined as derivative of $F_g$ with 2D viewpoints and ranked as given in Eq. (1):

$$[v_1, v_2, \ldots, v_n] = \left[ \frac{\partial F_g}{\partial I_1} \Big|_s, \frac{\partial F_g}{\partial I_2} \Big|_s, \ldots, \frac{\partial F_g}{\partial I_k} \Big|_s \right] \tag{1}$$

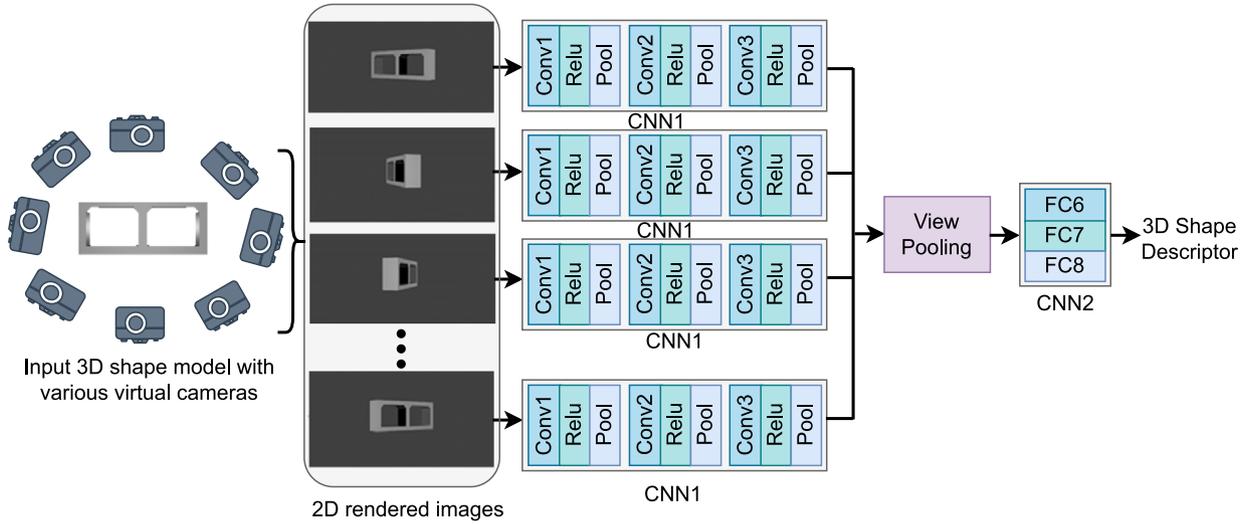

**Fig. 3.** Architecture of MVCNN.
*Source:* Adapted from Su et al. [35].

However, the view-pooling layer retains only the largest elements from each viewpoint, leading to some loss of information. A typical MVCNN architecture is illustrated in Fig. 3. [36] implemented MVCNN for classifying ten defects in road infrastructure. The author compared the classification performance between MVCNN and PointNet [37]. The results demonstrated the superior performance of MVCNN for CM of road infrastructure with mAcc of 98% compared to 83% in the case of PointNet. [38] proposed a CNN model to extract global features from regularly structured depth images. This approach contrasts with existing methods like MVCNN and PointNet, which utilize unstructured PC data. The depth images utilized in this study do not introduce any geometry loss, enabling fine-grid shape classification of defects in solder joints, which is crucial for CM of industrial systems. Group-View CNN (GVCNN) [5] introduces a hierarchical shape descriptor by incorporating grouping and individual viewpoints information in the pooling process. While GVCNN exhibits a significant improvement in accuracy compared to MVCNN, it faces challenges, particularly with smaller views, which can be a limitation for CM tasks. Multi-view harmonized bi-linear network (MHBN) [39] combines local convolutional features from multiple view using bi-linear pooling to generate a global shape descriptor. Later, the sequential behavior of the captured views was explored to recognize the 3D shapes. [40] combined CNNs and long short-term memory (LSTM) to aggregate multi-view features into shape descriptors. This approach leverages the temporal dependencies among views, enhancing the understanding of 3D shapes by fusing spatial and sequential information. SeqViews2SeqLabels [41] takes into account the spatial relationship among viewpoints by introducing an encoder to aggregate the information from sequential views and a decoder for predicting global features or sequence labels. Subsequently, the author extends this approach with 3D2SeqViews [42], efficiently aggregating information from both views and sequential spatial views in a hierarchical attention (view-level and class-level) mechanism. However, these methods are limited to aggregating ordered views and do not handle aggregating unordered views.

Another hierarchical network based on view graph representation was introduced in view-based graph convolutional network (view-GCN) [44]. Using this approach, the author constructed a view graph, where multiple views are treated as graph nodes. The view-GCN learns discriminative shape descriptors based on the relationship between multiple views. Based upon this concept, multi-view GCN (MVGCN) [43] was proposed for classifying defects (scratch, dent, protrusion) in synthetically generated 3D PC datasets on an aircraft fuselage. Here, a two-step analysis was performed: (a) an adaptive threshold method based on variation in local surface and (b) defect classification by the applicability of graph-based representations for capturing complex relationships among multiple views in the context of defect classification. Fig. 4 depicts the MVGCN method for identifying surface defects in automotive components. The threshold-based approach clusters all points into two groups, as defined in Eq. (2), which helps to identify potential defect regions. The graph-based representation is then utilized to classify the detected defects based on the relationships between multiple views.

$$f(x) = \begin{cases} 1 & \text{if } \sigma(x) \geqslant \delta \\ 0 & \text{if } \sigma(x) < \delta \end{cases} \tag{2}$$

where $\sigma$ represent local surface variation, $\delta$ is threshold, and $f(x)$ refers to center point $x$ in the point set $X$ representing defective regions ($f(x) = 1$) or normal region $f(x) = 0$. The second stage utilizes MVGCN to extract features and classify defects. It consists of a graph convolutional layer (GCL), multi-view GCL (MVGCL), and fully-connected (FC) layer.

To address the limitations of traditional pooling techniques in multi-view-based methods, recent approaches have explored more sophisticated fusion and attention mechanisms. Multi-view-based fusion pooling (MHFP) [45] adopts a hierarchical approach to fuse multi-view features into a compact descriptor, leveraging correlations between several views. This method effectively removes redundant information while retaining maximum relevant information using a 3D attention module to construct a graph. Similarly, multi-view softpool attention networks (MVMSAN) [46] refine view feature information using a soft-pool attention convolution framework. The attention mechanism plays a crucial role in addressing challenges related to down-sampling, feature information loss, and insufficient detail feature extraction, ultimately contributing to improved model performance. With the recent success in vision transformer (ViT) [11,12] proposed multi-view convolutional ViT (MVCVT), which combines CNN on each view to extract multi-scale local information and utilizes transformers to capture the relevance of multi-scale information across different views. This integration showcases the adaptability and effectiveness of transformer-based architectures in the context of multi-view feature extraction for 3D shape classification in CM applications.

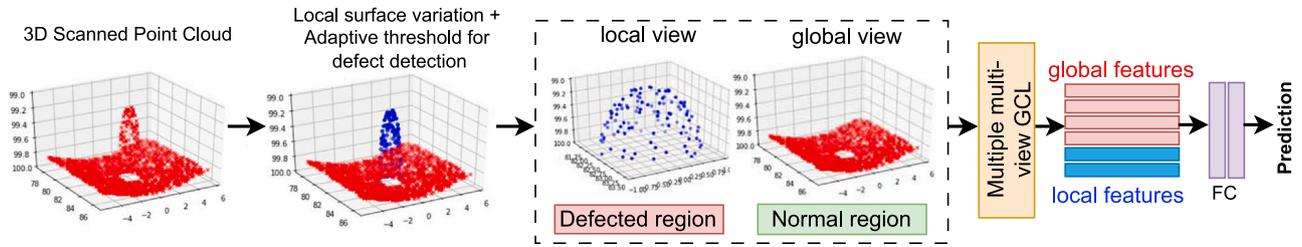

**Fig. 4.** Multi-view GCN for predicting surface defects in automotive parts.
*Source:* Adapted from Wang et al. [43].

In summary, view-based methods learn from view features to obtain global feature descriptors, leveraging established CNN frameworks. However, traditional pooled downsampling techniques can result in insufficient extraction of view refinement feature information, leading to a substantial loss of valuable insights from the view features. The more recent approaches, such as MHFP, MVMSAN, and MVCVT, have demonstrated the potential of advanced fusion and attention mechanisms to address these limitations and further enhance the performance of multi-view based methods for industrial CM applications.

### 3.1.2. Volumetric-based methods

This method represents 3D shapes in the form of a 3D voxel grid using 3D volumetric convolutions such that each voxel signifies whether a point in 3D space is occupied by an object or not. VoxNet [47] addresses large PC data by integrating a volumetric occupancy grid with 3D CNN. [26] proposed a convolutional deep belief network, 3D ShapeNets to represent a 3D shape based on the probability distribution (PD) of binary variables on voxel grids. However, these methods do not perform well in processing dense 3D data due to high computation and memory requirements for higher resolution (computational complexity is a cubic function of voxel grid resolution) [48]. To overcome this limitation, a hierarchical compact structure needs to be introduced. OctNet [49] achieves this by partitioning 3D PC data hierarchically using a set of unbalanced octrees, where each leaf node stores a pooled summary of the features of the voxels. This approach focuses memory allocations on the relevant regions, enabling the use of deeper networks with high resolution. Subsequently, Octree-based CNN (O-CNN) [6] was proposed for 3D shape classification. O-CNN averages the normal vectors of a 3D model into fine-leaf octants as network input and performs 3D CNN over the octants occupied by the 3D shape surface. Another network based on the non-uniform indexing named Kd-Net [50] was introduced to mimic the convolutional-based network. Kd-Net requires small memory and computation in comparison to uniform grids. [51] used 3D grids to represent PC data, further expressed using 3D modified Fisher vector method. This vector acts as an input to the 3D CNN to produce global features. To address the challenges of low-resolution voxels and high computation requirements, Point-Grid [52] introduced a hybrid network that integrates both the grid and point representation of the PC data. In [53], a multi-orientation volumetric DNN (MV-DNN) was proposed to limit the octree partition to a certain depth for reserving leaf octants with sparse features. This method improves classification for both low and high-resolution grids.

In summary, volumetric-based methods represent 3D PCs using voxel grids to address the data's unordered structure. However, this approach requires input voxels to be in a regular form for convolutional operations, leading to information loss with low-resolution voxels and subsequently lower classification accuracy. Additionally, these methods face challenges related to high computation requirements, especially for high-resolution data.

### 3.2. Direct point based methods

Direct point-based methods directly process the input PC data to produce a sparse representation. These methods extract a feature vector for each point by aggregating the features of neighboring points. In this way, models designed for raw PC data typically begin by extracting low-dimensional features from individual points and later aggregate them to obtain high-dimensional features. Direct point-based methods can be further categorized into point-wise multi-layer perceptron (MLP), convolution-based, and graph-based methods.

### 3.2.1. Pointwise MLP methods

These methods independently process each point in the 3D points through shared MLPs to extract local features. PointNet [37] model represents unordered PCs as a set of 3D points $\left\{ P_i \mid i = 1, 2, \ldots, n \right\}$. Here, $P_i$ is the vector consisting of $x, y, z$ coordinates along with feature channels such as color, normal vectors, etc. These local features are extracted independently for each point through multiple MLP layers and aggregated to obtain global features using a symmetric aggregation function on the transformed elements in the set, as given by Eq. (3).

$$f(x_1, x_2, \ldots, x_n) \approx g(h(x_1), h(x_2), \ldots, h(x_n)) \tag{3}$$

such that

$$\begin{cases} f : 2^{\mathbb{R}^N} \to \mathbb{R} \\ h : \mathbb{R}^N \to \mathbb{R}^K \\ g : \underbrace{\mathbb{R}^K \times \cdots \times \mathbb{R}^K}_{n} \to \mathbb{R} \end{cases} \tag{4}$$

is symmetric function. Here, $h$ and $g$ can be approximated using the MLP network and the composition of the single-variable and max pooling functions, respectively. This collection of $h$ function captures various properties of the $N$ 3D point set ($\mathbb{R}^N$) by learning the function $f$'s. The output from Eq. (3) forms a vector $[f_1, f_2, \ldots, f_K]$, which acts as a global feature of the input set. Later, the classifier is trained on the shape global features for classification. This approach allows PointNet to directly process unordered PC data without the need for converting it into a regular grid or multi-view representations, making it a versatile and efficient method for industrial CM tasks.

Following the success of PointNet, several extensions and improvements have been proposed to enhance the performance of pointwise MLP methods for 3D shape analysis and defect detection in industrial systems. For example, PointNet++ [54] introduced a hierarchical feature extraction process to capture local structures at multiple scales, addressing the limitations of the original PointNet in handling complex geometric structures often encountered in industrial applications. Fig. 5 provides a visual comparison of the architectures of the PointNet [37] and PointNet++ [54] models, respectively. Pointwise MLP methods, exemplified by PointNet, have shown great promise for industrial CM applications due to their ability to directly process unordered 3D PC data. For instance, [55] investigated the use of PointNet to detect defects (scaling, delaminations, and spalls) on the bridge surfaces. This work collected 55 parts of the scanned bridges containing 13.5 M points with 847 defects to obtain an mAP of 85.7% with 11.6% loss for the testing dataset. Building upon the success of PointNet, various models have been proposed in the literature for the direct processing of 3D PC data in industrial CM applications. [56] introduced a dual-level-defect detection PointNet ($D^3$PointNet) for inspecting defects in solder paste patterns in printers, using segmentation and multi-label classification. This approach was designed to address the challenges of sparseness and varying sizes of the solder patterns, where conventional CNNs may not be suitable. Therefore, defect detection is performed at two semantic levels: micro and macro, providing robustness to changes in sparsity and input data size. The author defined two hand-crafted features, edge and prior features, to prevent the loss of spatial information in the PC during processing. The work achieved a mAP of 97.87% and mPrec of 97.28%. Another example is the Self-organizing networks (SO-Net) [57], which achieves permutation invariance for unordered PCs by building a self-organizing map based on the spatial distribution of PCs. The hierarchical feature extraction of SO-Net results in a single feature vector that represents the entire PC. To enhance the performance of PointNet++, [58] proposed PointNeXt, introducing an inverted residual bottleneck design with separable MLPs into the PointNet++ architecture. This modification results in an effective and efficient model with a 10× faster inference, which is crucial for real-time CM applications.

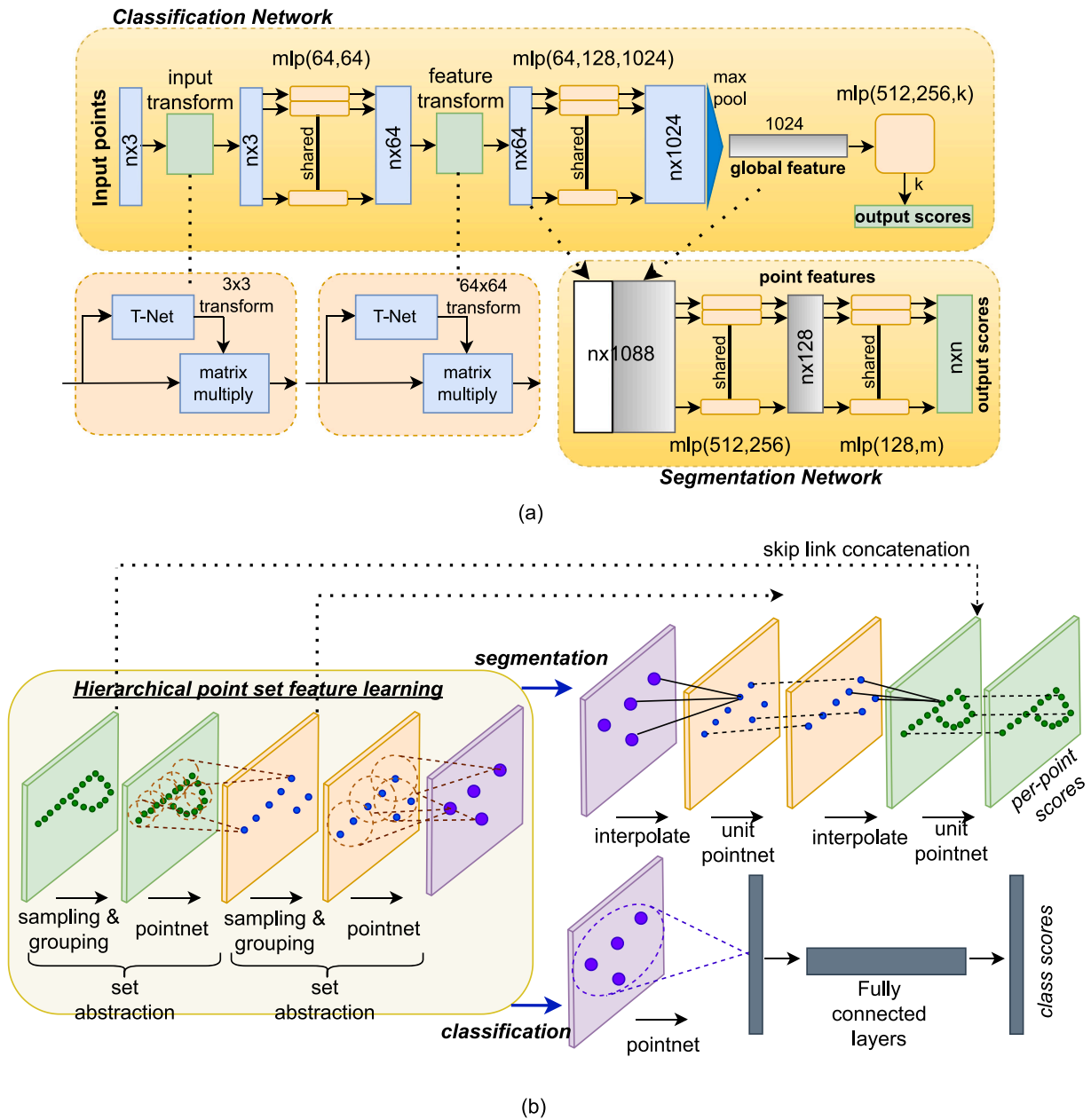

**Fig. 5.** Architecture comparison of state-of-the-art methods; (a) PointNet [37] and (b) PointNet++ [54] respectively.
*Source:* Adapted from Qi et al. [37,54].

In addition to the pointwise MLP methods, several networks in the literature have leveraged geometrical features for 3D PC processing in industrial CM applications. Based on PointNet [37], Motion-based network (MO-Net) [59] incorporates the context of 3D geometry in the form of a finite set of moments as network input. This approach uses an attention mechanism to learn fine-grained local features of the PC, which can be beneficial for capturing subtle defects in industrial systems. Similarly, Point attention transformers (PATs) [60] represent each point in the PC using its absolute and relative positions concerning its neighbors. Then, group shuffle attention (GSA) captures the relations between these points, and a differentiable, permutation invariant, and trainable end-to-end gumbel subset sampling (GSS) layer is developed to learn hierarchical features that can be important for effective defect classification and localization. PointNet++ [54] based networks, such as PointWeb [61], explore the interaction between points using an adaptive feature adjustment module. This module interconnects all point pairs in a local region, forming a fully linked web to describe local regions for 3D recognition in industrial applications. However, these methods require large sample sizes that might not be available in real manufacturing environments. To address this limitation, [62] introduced a tensor voting-based method is introduced for classifying surface anomalies such as debris patches, depression marks, slag, pinholes, and oscillation marks on steel surfaces. Tensor voting is employed to infer geometric characteristics such as curvature, surface, and junction by voting over the neighborhood to identify points that may contain potential anomalies. Aggregated descriptive features are extracted for each selected PC sample and fed into a sparse multiclass SVM classifier for anomaly classification and feature selection. The approach achieved an average accuracy of 86.27%. Another example is the Strictly rotation-invariant network (SRI-Net) [63], which projects the PC data into a rotationally invariant representation, utilizing a PointNet backbone to extract global features and a graph aggregation method to extract local features. This can be particularly useful for industrial applications where the orientation of the object may vary. In addition, PointASNL [64] adaptively adjusts the coordinates of the initially sampled points using the furthest point sampling (FPS) algorithm and introduces a local-on-local (L-NL) module to capture local and long-range dependencies of these sampled points. PointGMM [65] is a coarse-to-fine feature learning method subdividing the input point data into distinct groups using a hierarchical Gaussian mixture model (hGMM). This approach focuses on learning features of small and large regions, respectively. Here, the bottom GMM focuses on learning features of small regions, while the top GMM learns features of larger regions. More recently, [66] proposed a novel improved PointNet++ for classifying and segmenting the sewer pipes' defects of different shapes and sizes. The author improved the network structure by incorporating residual connection and cross-entropy loss with label smoothing in the network. Later, the training process is optimized by AdamW and the cosine learning rate decays. These geometrical feature-based methods demonstrate the versatility of 3D PC processing techniques for addressing the unique challenges of industrial CM applications, where the ability to effectively capture and leverage the underlying geometric properties of the data can be crucial for accurate defect detection and classification.

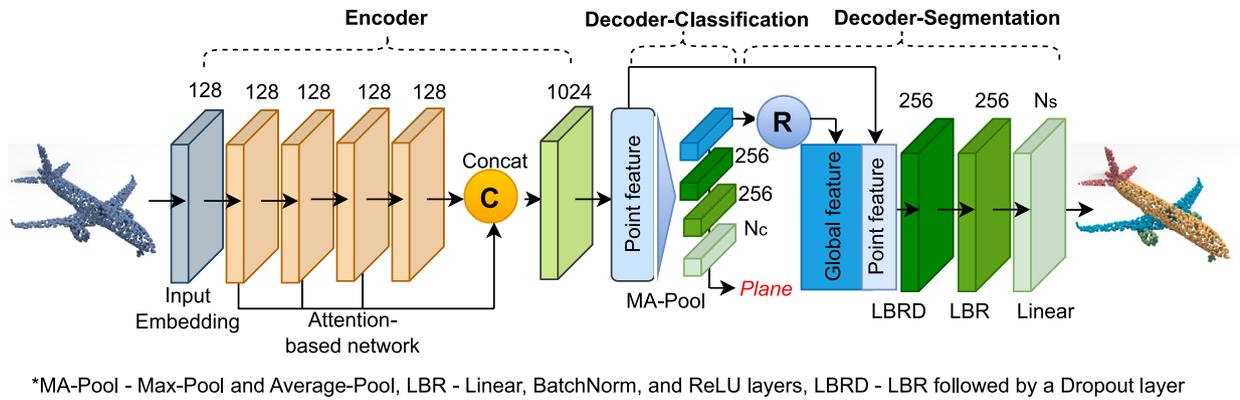

**Fig. 6.** Architecture of PCT [67]. The encoder layer consists of an input embedding module with a stacked attention module, while the decoder contains multiple Linear layers.
*Source:* Adapted from Guo et al. [67].

Unlike the above methods, PC transformer (PCT) [67] is a prominent example, which is based on a permutation-invariant transformer rather than a self-attention mechanism for handling unstructured and disordered point data with irregular domains. The overall architecture of PCT is presented in Fig. 6. PCT transforms (encodes) the raw PC into a new feature space to characterize the semantic affinities between points. These features are then fed into the attention module to learn the discriminative representation for each point, followed by a linear layer to generate the final output feature [72]. Meanwhile, 3DMedPT [13] proposed a transformer-based network for analyzing 3D medical (healthy and abnormal blood vessels) PC data. This paper compares 3DMedPT with PCT, DGCNN, and SO-Net to achieve a high mAP of 94.06% and F1-score of 93.6, respectively. Expanding on this, a Transformer-based network (TR-Net) [73] utilizes a neighborhood embedding strategy and residual backbone with skip connections to enhance context-aware and spatial-aware features. The author uses an offset attention operator on PC spatial information to sharpen the attention weights to improve the extraction of global features for CM tasks. Inspired by the bi-directional encoder in transformers (BERT), Point-BERT [74] adopts a strategy of dividing the PC into distinct local blocks, generating discrete point labels that represent local information using a PC marker. This approach allows the model to capture specific details and features within localized regions of the PC. Similar to BERT, Point-BERT introduces a masking mechanism where some input PCs are randomly masked and then fed to the backbone transformer network. This facilitates bidirectional learning and enhances the model's ability to capture contextual relationships in 3D PC data.

However, the uneven distribution of information in the PC may lead to a loss of information during the reconstruction task. To address this challenge, masked autoencoder (MAE) methods, such as Point-MAE [75], have been proposed. Point-MAE is a self-supervised learning (SSL) method designed to mitigate issues related to uneven information density and information leakage of PC locations. In a study by [70], a deep autoencoder network was proposed for processing 3D PC data of concrete bridges, which are critical industrial infrastructure. The network takes encoded shape and neighborhood features as inputs and uses a one-class support vector machine (OC-SVM) to classify spall defects on the concrete bridge's PC data. The author tested the network on a diverse set of quasi-real PCs covering a variety of (low, medium, and high) noise and defect conditions, achieving a mAcc of 98% and an F1 score of 69% for CM tasks. Expanding on transformer-based approaches, PointConT [76] presents a novel approach to 3D PC processing by leveraging transformer-based clustering and self-attention mechanisms. The method focuses on clustering points based on their content and applying self-attention within each cluster. This design aims to capture long-range dependencies within the PC while managing computational efficiency. Additionally, the authors introduce an inception feature aggregation module featuring a parallel structure to aggregate high and low-frequency information separately. [77] integrates clustering and multi-class classification within in-site PC processing for surface defect detection in additive manufacturing (AM) industries. Eight classifiers – Support Vector Machine (SVM), K-Nearest Neighbors (K-NN), Gaussian process (GP), Decision Tree (DT), Naïve

**Table 2**
Performance evaluation for classification methods on industrial applications.

| Ref. | Application | Classes | Method | Results | Points/Objects |
|------|-------------|---------|--------|---------|----------------|
| [68] | Defect classification in sewer | 4 classes: normal, displacement, brick, and rubber ring | DGCNN | OA = 47.9 | 17,027 |
| | | | | mIoU = 46.1 | |
| | | | PointNet | OA = 18.4 | |
| | | | | mIoU = 18.5 | |
| [36] | Classification of infrastructure elements | 10 classes: column, 3 types of culverts, 5 types of walls, and sump | PointNet | Mean F1 score = 89.3 | 1496 |
| | | | PointNet | OA = 83 | |
| | | | | F1 score = 87 | |
| [62] | Classification of anomalies on steel surfaces | 5 classes: debris, oscillation, slag, depressions, and pinholes marks | Tensor voting | Mean Acc = 86.27 | 96,266 |
| [43] | Defect classification in precast concrete specimen | 2 classes: defective, and normal | MVGCN | Euclidean = 97.9 | 2000 |
| | | | | Geodesic = 93.8 | |
| | | | DGCNN | Euclidean = 70.8 | |
| | | | | Geodesic = 81.3 | |
| [66] | Defect classification in concrete sewer pipes | 5 classes: 3 circular defects of varying diameter, square and triangular defect | Improved PointNet++ | Mean F1 score = 68.15 | 1.4 M |
| | | | | Accuracy = 73.01 | |
| | | | PointNet++ | Mean F1 score = 61.36 | |
| | | | | Accuracy = 67.55 | |
| [69] | Defect classification in polyvinyl chloride-sewer pipes | 4 classes: normal, and defective (brick, rubber ring, displacement) | TransPCNet | F1 score = 60.58 | 17,027 |
| | | | | Precision = 61.47 | |
| | | | DGCNN | F1 score = 16.66 | |
| | | | | Precision = 34.55 | |
| | | | PointNet | F1 score = 30.23 | |
| | | | | Precision = 28.61 | |
| [55] | Classification of surface defects on bridges | 4 classes: cracks, spalling, scaling, and delaminations | PointNet | mAcc = 85.7 | 21 M |
| [70] | Spall Classification on bridges | 2 classes: normal and defective | Point-wise | mAcc = 98 | 21 M |
| | | | | Precision = 68 | |
| | | | PointNet | mAcc = 97 | |
| | | | | Precision = 61 | |
| [56] | Classification of printer defects | 2 classes: normal and defective solder patterns | D3PointNet | mAcc = 97.17 | 4.2 M |
| | | | | Precision = 97.28 | |
| [38] | Classification of solder joints shapes | 2 classes: normal and defective | SDCNN | mAcc = 98.1 | 800 |
| | | | | Precision = 83.9 | |
| | | | MVCNN | mAcc = 93.6 | |
| | | | | Precision = 76.9 | |
| [71] | Classification of concrete sewer pipes | 2 classes: potholes, and background | Improved PointNet++ | mAcc = 73.01 | 17,027 |
| | | | PointNet ++ | mAcc = 67.55 | |

Bayes (NB), Artificial Neural Networks (ANN), Random Forest (RF), and AdaBoost (AB) – were examined and fine-tuned. The KNN model exhibited superior performance, achieving a remarkable accuracy of 93.15% for this industrial application. Similarly, [78] leverages macro-level data on neighboring points within the PC through a patch-based strategy for detecting defects in the AM sector. Five machine learning (ML) techniques – Bagging of Trees (BoT), Gradient Boosting (GB), RF, K-NN, and SVM – were evaluated under diverse operational settings. Bagging and RF emerged as the top-performing models for defect identification, achieving accuracies of 99.99% and 99.59%, respectively, with a patch size of 20. Most recently, [69] proposed a transformer-based PC classification network (TransPCNet) for CM of sewer pipelines. TransPCNet comprises a feature embedding module for extracting features from local neighbors, an attention module designed to learn and enhance feature extraction, and a classification module. Additionally, the authors introduced a weighted smoothing cross-entropy loss to aid the network in feature learning while addressing PC imbalances. These advanced techniques, ranging from masked autoencoders and deep autoencoders to transformer-based methods and hybrid ML approaches, demonstrate the continued evolution of 3D PC processing for industrial CM applications.

In summary, pointwise MLP methods demonstrate efficiency and effectiveness in processing raw 3D PC data, leveraging simplicity to capture local features independently for each point, allowing for fine geometric structure understanding. This makes them well-suited for industrial CM applications, where the input data may be irregularly distributed and difficult to represent using regular structures. Despite their advantages, challenges arise when handling large-scale and complex PCs due to limitations in capturing long-range dependencies and holistic context. Additionally, these methods face difficulties in accommodating variations in point density, leading to potential impacts on the robustness of feature extraction. The introduction of point-based transformers and related models addresses some of these challenges by leveraging permutation-invariant transformers. These transformer-based approaches excel in managing unstructured and disordered point data, presenting a promising avenue for advancing the processing of 3D PC data.

### 3.2.2. Convolution-based methods

Following the remarkable success of CNNs in CV tasks [79,80] such as image classification, object detection, and segmentation, there has been a significant effort to extend these methodologies to analyze geometric and spatial data. Unlike the regular grid structure in 2D images, geometric data (PCs, 3D models, etc.) lacks underlying grid information, necessitating the development of new methods. Several convolution-based methods, including *continuous* and *discrete convolution-based methods*, have been developed for analyzing 3D PC data in industrial CM applications [81–83]. 3D continuous convolution methods are defined in a continuous space where weights for neighboring points are spatially related to their center point. Conversely, 3D discrete convolutions involve a fixed-size kernel sliding over a structured point grid, with weights assigned to neighboring points determined by their offsets relative to the center point of the kernel.

Among **continuous convolution methods**, PointConv [84] stands out by representing convolution kernels as non-linear functions of the local coordinates of 3D points. These functions comprise weight (learned with MLP layers) and density (learned by kernel density estimation) functions as represented in Eqs. (5) and (6), respectively. PointConv efficiently computes the weight function, providing translation and permutation-invariant convolution in 3D space. It is known that continuous 3D convolution can be expressed as:

$$Conv(W, F)_{xyz} = \iiint_{(\delta_x, \delta_y, \delta_z) \in G} W(\delta_x, \delta_y, \delta_z) F(x + \delta_x, y + \delta_y, z + \delta_z) d\delta_x \delta_y \delta_z \tag{5}$$

where $F(x + \delta_x, y + \delta_y, z + \delta_z)$ represents the point feature and $W(\delta_x, \delta_y, \delta_z)$ is the weight function approximated by MLPs in local region $G$ with center point $p = (x, y, z)$ for any random position $(\delta_x, \delta_y, \delta_z)$. Based on this, PointConv ($PConv$) can be defined as:

$$PConv(S, W, F)_{xyz} = \sum_{(\delta_x, \delta_y, \delta_z) \in G} S(\delta_x, \delta_y, \delta_z) W(\delta_x, \delta_y, \delta_z)$$
$$\times F(x + \delta_x, y + \delta_y, z + \delta_z) \tag{6}$$

where $S(\delta_x, \delta_y, \delta_z)$ is inverse density scale at any position $(\delta_x, \delta_y, \delta_z)$ calculated using kernel density estimation (KDE) followed by 1D non-linear transform fed with MLPs. This $S$ needs to be estimated due to the non-uniform PC input data.

Another notable method, KPConv [85], introduces a deformable convolution operator that learns local shifts at each convolution location, enabling adaptation of the kernel shape based on the input PC's geometry. This adaptive capability is particularly valuable for handling the complex and irregular structures often encountered in industrial PC data. Another approach, ConvPoint [86] takes a different approach by introducing a dense weighing function to define detailed and adaptive convolutional kernels. In this method, the derived kernel is explicitly represented by a set of points, each associated with specific weights, allowing for a more flexible and customized convolution operation for industrial CM tasks.

In the realm of **discrete convolution-based methods**, PointCNN [87] is a pioneering work that tackles the unordered and irregular structure of 3D PC data. PointCNN learns $\chi$-transformation from input PC data to *project* or *aggregate* information into few representative points ($9 \rightarrow 5 \rightarrow 4$) while preserving rich feature information, effectively addressing the challenges posed by the inherent structure of PC data obtained from industrial sensors and scanners. This transformation permutates the weight of input point features into canonical order. A CNN is then applied to these transformed features, addressing the unordered and irregular structure of the 3D PC data. Additionally, Pointwise CNN [88] applies the convolution operator on each point in the PC to learn pointwise features. Here, varying neighboring points lying within each kernel contribute to the center point in each convolution layer, given by:

$$P_i^l = \sum_k W_k \frac{1}{|\varphi_i(k)|} \sum_{x_j \in \phi_i(k)} P_j^{l-1} \tag{7}$$

where $W_k$ is kernel weight while $\varphi_i(k)$ is $k$th sub-domain of the kernel centered at point $i$. $x_i$ is the coordinate of point $i$ while $P_i$ and $P_j$ value of point $i$ and $j$, and $l-1$ and $l$ are index of input and output layer respectively. The obtained outputs are then concatenated before being fed to the final convolution layers for segmentation or fully connected layers for object recognition. This approach can be particularly useful for extracting localized features in industrial CM applications, where the spatial relationships and interactions between individual points in the PC may hold valuable insights. Unlike the traditional methods, Pointwise CNN does not require up-sampling or down-sampling of the PCs. Based on PointCNN, spherical harmonics network (SPH-Net) [89] proposed a rotation invariance CNN on PCs by using spherical harmonics-based kernels at different network layers. SC-CNN [90] implements a spatial coverage convolution by constructing an anisotropic spatial geometry in the local PC and replacing the depthwise convolution with the spatial coverage operator (SCOP). This method excels in learning high-order relations between points, providing shape information, and enhancing network robustness.

In summary, continuous convolution methods, such as PointConv, KPConv, and ConvPoint, offer adaptability to diverse PC geometries and effective pattern capture. However, they overlook PC distribution considerations. In the discrete domain, PointCNN efficiently handles unordered structures, Pointwise CNN excels in pointwise feature learning, SPH-Net introduces rotation invariance, and SC-CNN learns high-order relations. While these methods enhance 3D PC analysis, challenges persist in distribution awareness and computational efficiency.

### 3.2.3. Graph based methods

Graph-based methods provide an alternative to CNNs for handling unstructured and unordered 3D PC data. Unlike CNNs, which operate on regular grid data, graph-based methods transform the PC into a comprehensive graph, avoiding the need for voxelization. A typical architecture of a graph-based PC network is illustrated in Fig. 7. This approach allows for flexibility in capturing intricate relationships among points, representing each point in the PC as a vertex in the graph, with edges established between nearby points [91]. These edges analyze spatial relationships, creating a graph that encapsulates the geometric features of the original PC. Graph-CNN [92], also known as PointGCN, classifies 3D PCs by combining localized graph convolution layers with two types of data-specific pooling layers (down-sampling). This method effectively incorporates the geometric information encoded in the graph, enhancing the robustness of the model. In contrast, Dynamic graph CNN (DGCNN) [8], inspired by PointNet, addresses the limitation of processing each point independently, as in PointNet, leading to the neglect of local features between points. To solve this, Dynamic CNN uses the EdgeConv layer to capture edge features from each point and its neighbors. EdgeConv explicitly constructs a local graph while learning the embeddings for the edges, enabling the grouping of the points in Euclidean and semantic space. [68] investigated the application of DGCNN and PointNet for classifying defects on synthetic and real sewer PC data. The author observed that the DGCNN network consistently outperforms the PointNet network for synthetic and real datasets. Dynamic points agglomeration module (DPAM) [93] is based on graph convolution to agglomerate (sampling, grouping, and pooling) points by multiplying the agglomeration matrix and points feature matrix. Based on PointNet and PointNet++, a hierarchical network is constructed by stacking multiple DPAMs by dynamically exploiting the relation between points and agglomerated points in a semantic space. Additionally, a variation of DGCNN, linked-DGCNN [94] simplifies the model by removing the transformation layer in DGCNN. This is implemented by connecting the hierarchical features of various dynamic graphs to address the issue of gradient vanishing. PointView-GCN [95] introduces a multi-level GCN to hierarchically aggregate shape features of single-view PCs. This method allows encoding both object geometric cues and their multiview relationships, improving the extraction of global features. Gaussian super vector network (GSV-NET) [96] is a recent approach that captures and aggregates both local and global features of the 3D PC to enhance the information of the PC features. GSV-NET combines the GSV network and a 3D-wide inception CNN architecture to extract global features. The method then converts 3D PC regions into color representations and employs a 2D-wide inception network to obtain local features. Also, [97] integrated the distance and direction information in GCN (DDGCN) by constructing a dynamic neighborhood graph. This dynamic graph utilizes MLPs and the similarity matrix to capture the local features of the PC. Additionally, the author modifies the loss function by incorporating the center loss, enhancing the discriminative power of the model.

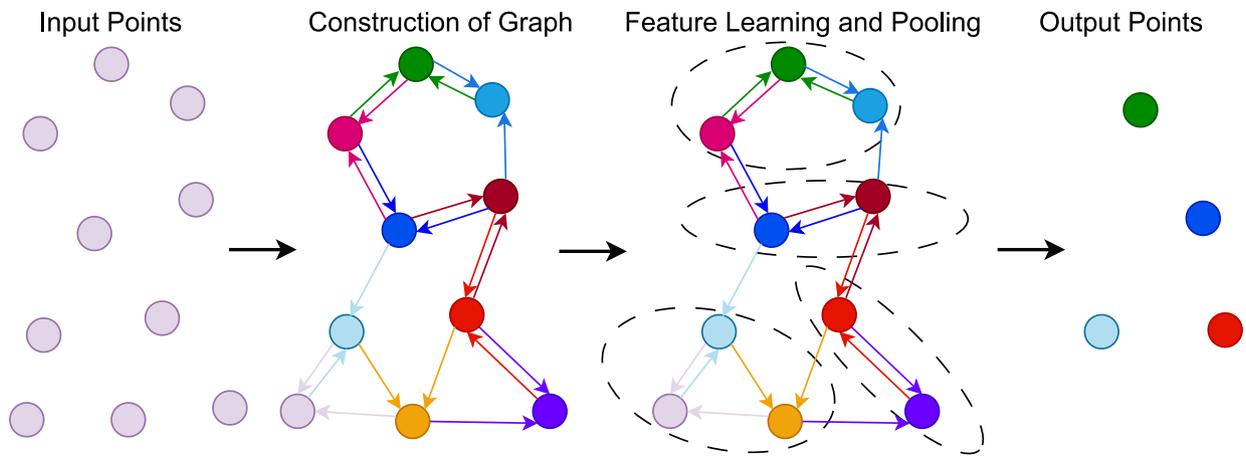

**Fig. 7.** An illustration of a graph-based 3D PC network.

Overall, point-based methods operate directly on raw PC data, rendering them well-suited for irregularly sampled and unstructured datasets with lower computational demands. Pointwise methods leverage MLP networks as fundamental building blocks for learning pointwise features, showcasing versatility in various network architectures. While literature indicates the superior performance of convolution-based networks for irregular PC data, limited research exists on continuous and discrete convolution networks in this context. Graph-based approaches provide another avenue for handling irregular PC data, but extending these methods, particularly those based on spectral domain graph structures, to various graph configurations remains a challenging task. Future research directions may explore advancing convolutional and graph-based methodologies to enhance point-based methods understanding and processing capabilities for diverse and complex 3D industrial datasets. Investigating methods that can effectively account for the underlying distribution of the PC data may be crucial for developing robust and reliable feature extraction and defect detection techniques in industrial CM applications. Table 2 present the outcomes of defect shape classification for industrial systems.

## 4. Deep learning for 3D PC segmentation

The effective segmentation of 3D PC data is crucial for a wide range of industrial CM applications. By accurately identifying and classifying the various elements within the PCs, industrial systems can benefit from enhanced defect detection, component-level analysis, and improved predictive maintenance capabilities. The task of 3D PC segmentation demands a comprehensive understanding of each point's geometric structure and intricate details in the 3D PC data. The segmentation task can be broadly categorized into three major types:

1. Semantic segmentation (Scene level): This method classifies each point within a 3D PC into predefined categories by assigning semantic labels based on their characteristics, enabling a high-level understanding of the overall industrial scene or environment.
2. Instance segmentation (Object level): This method identifies and distinguishes each object in the 3D PC by assigning each point with a specific instance or object. Unlike semantic segmentation, which groups points into predefined categories, object level segmentation enables the recognition of separate instances of objects, even if they belong to the same semantic class. This can be particularly valuable for industrial asset tracking, monitoring, and maintenance.

3. Part segmentation (Part level): This method segments each component of the object in the 3D PC providing a more detailed object-level segmentation. Unlike semantic segmentation, which categorizes points into high-level classes, and instance segmentation, which identifies and distinguishes individual objects, part segmentation provides a more detailed breakdown of each object by segmenting its constituent parts. This can be beneficial for component-level analysis, defect detection, and predictive maintenance of industrial equipment and systems.

These segmentation categories address different levels of abstraction, ranging from scene-level context to object-level identification and even detailed part-level segmentation. The ability to accurately perform these types of segmentation tasks on industrial 3D PC data can greatly enhance the understanding and analysis of complex industrial systems, leading to improved CM, reliability, and overall performance. The annotated examples for semantic, instance, and part segmentation on benchmark datasets are shown in Fig. 8.

### 4.1. 3D semantic segmentation

3D semantic segmentation, a key aspect of scene understanding, involves categorizing points in a 3D PC into predefined classes or labels. Similar to 3D shape classification, semantic segmentation methods can be divided into the following categories: projection-based methods (multi-view representation, spherical representation, and volumetric representation), direct point-based methods (pointwise MLP methods, convolution-based methods, and graph-based methods) [2,98–100]. Fig. 9 illustrates the most recent methods in 3D semantic segmentation.

#### 4.1.1. Projection-based methods

Projection-based methods in 3D semantic segmentation transform 3D PC data into 2D images. This transformation is achieved through various techniques, including multiview, volumetric, and spherical projections.

**Multi-View Representations:** Multi-view representation leverages the projection of 3D PCs into 2D images from multiple viewpoints for semantic segmentation. Researchers have explored various multi-view representation methods for industrial PC processing. [101] projected PCs into 2D images using multiple camera views and then processed by fully convolutional networks (FCNs) for semantic segmentation. The resulting pixel-wise segmentation was re-projected into the original input PC. The final semantic label for each point is obtained by fusing the re-projected scores over the different views. However, there is a loss of information during the projection process. To address this limitation, [102] pre-process the input images by generating a mesh and then decimates the PC to get a lighter cloud by voxelizing the scene.

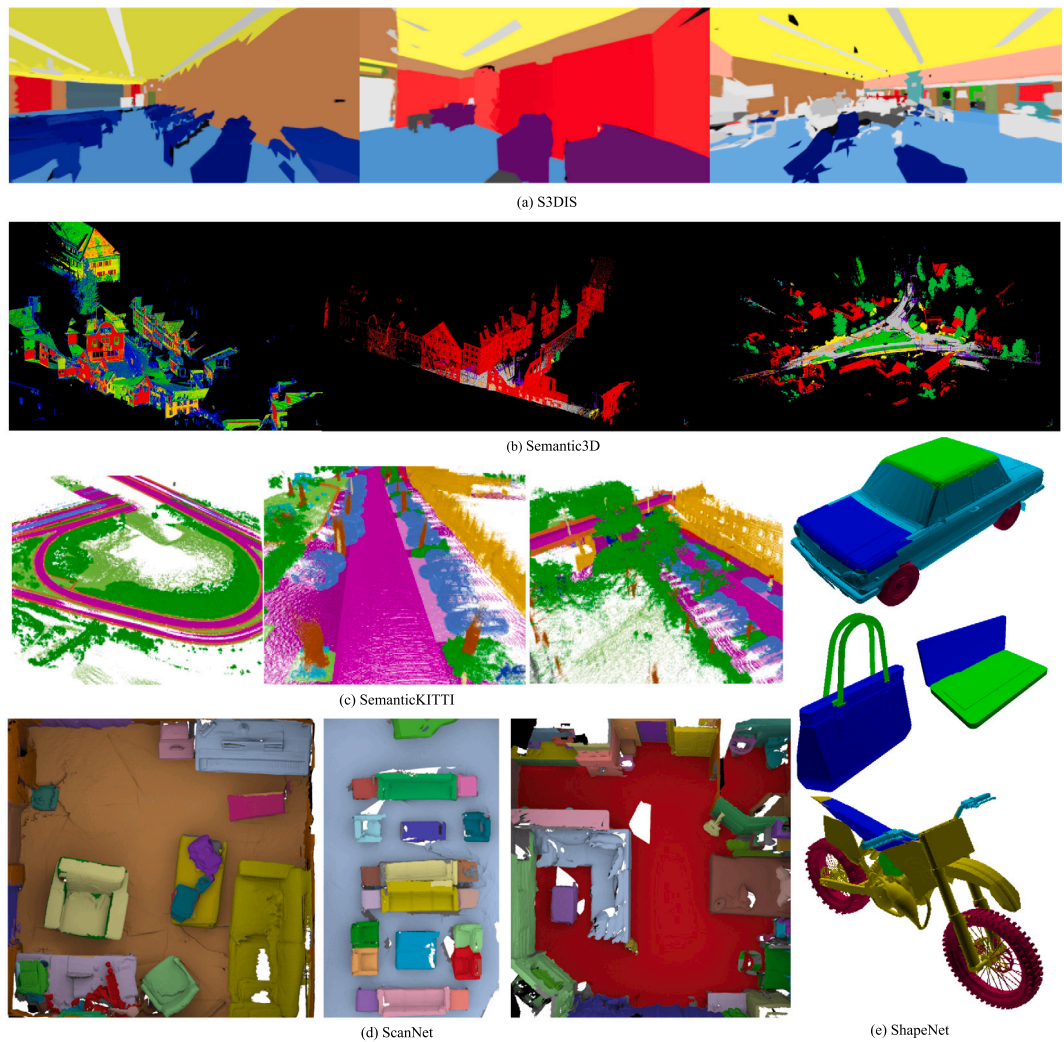

**Fig. 8.** Annotated examples for (a) S3DIS, adapted from Armeni et al. [19], (b) Semantic3D, adapted from Hackel et al. [20], (c) SemanticKITTI, adapted from Geiger et al. [22] for 3D semantic segmentation, (d) ScanNet, adapted from Dai et al. [18] for 3D-instance segmentation, and (e) ShapeNet, adapted from Chang et al. [27] for 3D-part segmentation.

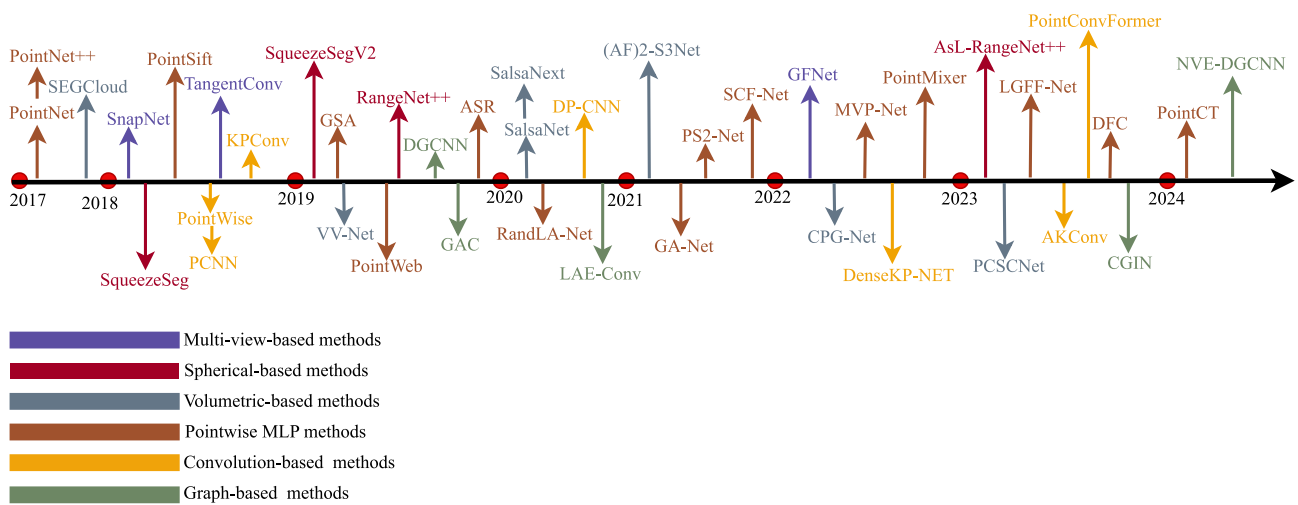

**Fig. 9.** Chronological overview of the most relevant DL-based 3D semantic segmentation methods.

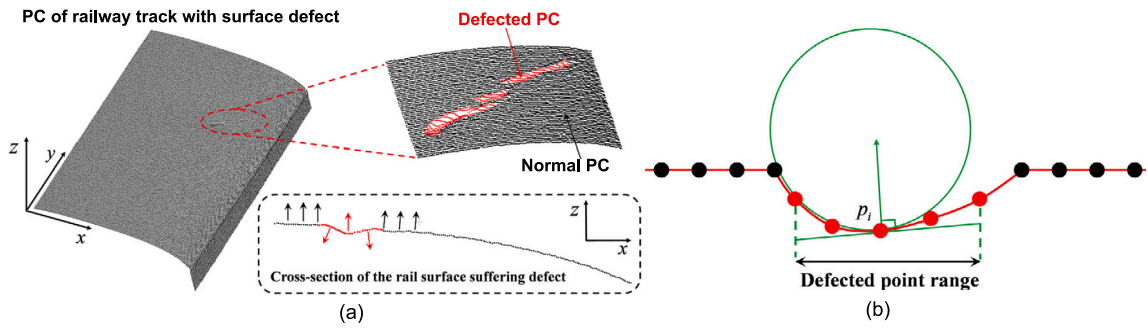

**Fig. 10.** Estimation of tensors on rail surfaces (a) Normals of defected region, (b) Curvature region.
*Source:* Adapted from Wang et al. [109].

Later, both color and volumetric information were used for semantic labeling. Multiple RGB and depth (RGB-D) images were generated to extract two generic features (normal deviation to vertical and noise estimation) using various camera positions to achieve this. The normal deviation to vertical at any point $p$ is given by:

$$NDev_p = arccos(|n_p.v|) \qquad (8)$$

where $n_p$ and $v$ are normal and vertical vector. The noise feature at any point $p$ is estimated based on the point spread in its neighborhood given by:

$$Noise_p = \frac{\lambda_2}{\lambda_0} \qquad (9)$$

where $\lambda_0$ and $\lambda_2$ are the highest and lowest singular values in PCA. Pixel-wise labeling is then performed on these captured snapshots and fed to two symmetrical encoder–decoder networks, such as Seg-Net [103] and U-Net [104], with a final fusion step using residual correction [105] on the obtained predicted scores. In a similar work, SnapNet [106] selected specific snapshots of the PC to generate pairs of RGB-D images. Then, pixel-wise labeling is performed on these 2D snapshots using FCNs. Extending this concept, [107] proposed a novel tangent convolution to design U-Net for segmenting dense PCs. This method involves projecting local surface geometry onto a virtual tangent plane, serving as input for subsequent tangent convolutions. Here, each tangent image can be treated as a regular 2D grid that supports planar convolution. Generic flow network (GF-Net) [108] proposes a novel approach for learning geometric features by fusing information from multi-view representations. The author used KNN post-processing over KPConv to make it end-to-end trainable. In a recent work by [109], a fusion of image and tensor-based PC, referred to as T-PCIF, was introduced for detecting defects on rail track surfaces. The approach employed YOLOv8 for swift defect localization and PC for precise damage detection. Moreover, image fusion was conducted through a combination of Kd-tree and superpixel segmentation algorithms. The rail surface was represented as 3D tensors, encapsulating rail points, neighboring points, and their associated features. Normal and curvature features were estimated to generate a feature tensor for identifying rail damages as illustrated in Fig. 10. The method utilized a total of 618 images, achieving a mAcc of 86.27% and mIoU of 70.18%.

Overall, multi-view representation methods project 3D PCs into 2D images from various viewpoints for semantic segmentation. While providing diverse perspectives, they are sensitive to occlusions and viewpoint selection, impacting performance. Tangent Convolution addresses geometric information by projecting local surface features onto a virtual tangent plane. However, these methods may not fully exploit inherent 3D geometric information, leading to potential information loss.

**Spherical Representations:** Spherical representation techniques leverage the projection of 3D PCs onto a spherical surface for efficient processing and segmentation. Researchers have explored various spherical representation methods for industrial PC processing. Based on CNN, an end-to-end pipeline named SqueezeSeg [110] was proposed to provide labeled point-wise output data. This method utilizes a conditional random field (CRF) as a recurrent layer to refine the segmented points further. To reduce the impact of dropout noise on the accuracy of SqueezeSeg, the author proposed SqueezeSegV2 [111]. SqueezeSeg2 introduced a novel CNN module named context aggregation module (CAM) to aggregate contextual information from a large receptive field, improving the network's robustness to dropout noise. However, challenges persist in handling issues arising from intermediate representations, including blurry CNN outputs and discretization errors.

Building upon these advancements, RangeNet++ [112] overcame the limitations of the previous methods by performing segmentation using CNN and an encoder–decoder hourglass-shaped architecture. The decoder in RangeNet++ incorporates a modified DarkNet [113] backbone architecture, enabling the use of aspect ratios beyond square configurations. Furthermore, RangeNet++ substitutes the CRF utilized in [110,111] with GPU-based nearest neighbor calculations across the complete PC. However, when dealing with unbalanced training samples, the training outcomes may become skewed, leading to inaccuracies in segmentation results. To improve the segmentation accuracy, AsL-RangeNet++, an extension of RangeNet++, introduces an asymmetric loss (AsL) function proposed by [114]. This method uses the AsL function with Adam optimizer, integrating asymmetric loss and probability transfer for calculating and adjusting object weights, enhancing the precision of semantic segmentation given by:

$$AsL = \begin{cases} L_+ = (1-B)^{\gamma_+} log(B) \\ L_- = (B_m)^{\gamma_-} log(1-B_m) \end{cases} \qquad (10)$$

such that

$$B = \sigma(z); \qquad B_m = \begin{cases} B - m, & B > m \\ 0, & B \leqslant m \end{cases} \qquad (11)$$

where $B$ is network probability while $m$ is marginal probability, and $\sigma(z)$ is sigmoid function. $L_+$ and $L_-$ are loss components of positive and negative samples based on the loss parameter $\gamma$. Compared to the loss function in RangeNet++, AsL accurately segments the PC in small proportions, and then, according to the loss parameters, it can adapt to the PC of different sample categories, improving the semantic segmentation. However, these methods stack point data from various modalities, such as coordinate, depth, and intensity, as inputs without accounting for their heterogeneous distributions.

**Volumetric Representations:** These methods transform unstructured 3D PCs into regular volumetric occupancy grids. The feature learning is then performed using NN to achieve semantic segmentation [27,47,115]. [116] projects the PC into occupancy voxels and fed into 3D-CNN to produce voxel-level labels, where all points within each voxel are assigned the same semantic label. This approach can be beneficial for tasks such as defect detection and component-level analysis in industrial systems. Building upon this, [117] introduced

InspectionNet, a 3D CNN-based framework designed to detect defects in synthetically generated concrete columns, demonstrating the potential of volumetric representations for infrastructure monitoring applications. SEGCloud [118] presented an end-to-end framework for semantic segmentation that integrates NNs, tri-linear interpolation (TI), and fully connected CRF (FC-CRF). This approach generates coarse voxel predictions using 3D-CNN, which are then transferred back to the raw input 3D points through TI. Finally, FC-CRF is used to enforce global consistency and improve the semantic understanding of the points, resulting in fine-grained segmentation results. Voxel variational autoencoder network (VV-Net) [119] used a combination of variational autoencoder (VAE) and 3D-CNNs to capture the point distribution within each voxel for semantic segmentation tasks, potentially enhancing the understanding of the underlying PC structure for industrial applications. [120] introduced SalsaNet, an encoder–decoder network comprising a series of ResNet blocks in the encoder and employing upsampling and feature fusion in the decoder. Subsequently, SalsaNext [121] enhanced SalsaNet by replacing the ResNet encoder with a stack of residual dilated convolutions and a pixel-shuffle layer in the decoder, facilitating uncertainty-aware semantic segmentation. (AF)2-S3Net [122], an extension of S3Net [123] and S3CNet [124], is an encoder–decoder model designed for 3D semantic segmentation using sparse-CNN. In this approach, the encoder incorporates an attentive feature fusion module to capture global and local features, while the decoder uses an adaptive feature selection module and feature map re-weighting to emphasize contextual information obtained from the feature fusion module. Cascade point-grid fusion network (CPG-Net) [125] adopts a cascading approach to extract and aggregate semantic features from point-view, bird's-eye view, and range-view representations. To improve robustness, a transformation consistency loss based on test-time augmentation is introduced to ensure agreement between original and augmented PCs. PCSC-Net [126] combines point convolution and 3D sparse convolution for semantic segmentation. It generates large-size voxels from input PCs, applies point convolution to extract voxel features, and then utilizes 3D sparse convolution to propagate features into neighboring regions, enhancing feature extraction and context understanding. However, volumetric methods may lose information with low-resolution 3D grids, and their computational costs and memory requirements increase cubically with voxel resolution.

While volumetric methods have proven effective in various PC processing tasks, they may suffer from information loss due to low-resolution 3D grids. Also, their computational costs and memory requirements increase cubically with voxel resolution. This can be a critical consideration for industrial CM applications, where real-time processing and efficient resource utilization are crucial.

### 4.1.2. Direct point-based methods

These methods operate directly on unstructured and irregular PCs, which poses a challenge for applying standard CNNs. PointNet [37] is a pioneering work in this domain, introducing a framework for processing direct PCs. Building upon PointNet, various approaches have been proposed, including pointwise MLP, point convolution, and graph-based methods, all aiming to enhance the processing and understanding of unstructured and unordered PC data.

**Pointwise MLP methods:** These methods utilize shared MLPs as the fundamental building block in their networks. However, the features extracted on a pointwise basis by these shared MLPs may face challenges in capturing the complex local geometry within PCs and the mutual interactions between points. To address these limitations, novel strategies have been introduced, including neighboring feature pooling, attention-based aggregation, and local–global feature concatenation.

*Neighboring feature pooling:* These methods are designed to capture local geometric patterns by aggregating information from nearby points to learn features for individual points in a PC. In [127], Point-Net [37] was employed for semantic segmentation of elements such as pipes, valves, and background in two underwater environments:

pool and sea. Additionally, the author created a novel PC dataset containing pipes and valve elements in various underwater scenarios. This work achieved an F1-score of 97.2% and 89.3% for the pool and sea test sets, respectively. PointNet++ [54] performs a hierarchical grouping of points to learn features from large local regions. Subsequent developments, such as multi-scale grouping and multi-resolution grouping, have been introduced to address challenges arising from the non-uniform density of PCs, well-suited for handling the non-uniform density often encountered in industrial PC data [128] proposed the surface-normal enhanced PointNet++ (SNEPointNet++) for infrastructure inspection and monitoring by semantic segmentation of defects, such as cracks and spalls, on concrete bridge surfaces. The method was tested on their custom dataset containing concrete surface defects with 1785 cracks and 2319 spalls with minimum width of 2 mm and 5 mm. This approach emphasizes utilizing normal vector, color, and depth characteristics to address challenges associated with small size and imbalanced PC data. If $S(x, y, z) = 0$ is any surface based on local surface fitting, then for point $P_0 = (x_0, y_0, z_0)$ on the surface, the normal vector $\bar{n}$ can be calculated by:

$$\bar{n} = \frac{\nabla S}{|\nabla S|} \tag{12}$$

where $\nabla S$ is gradient of $S$, and $|\nabla S|$ is vector length. Furthermore, based on PointNet++, normalized coordinates $(X_i, Y_i, Z_i)$ are used to provide the relative information of each point in a segment expressed as:

$$X_i = \frac{x_i}{x_{max}}, Y_i = \frac{y_i}{y_{max}}, Z_i = \frac{z_i}{z_{max}} \tag{13}$$

where $x_{max}, y_{max}, z_{max}$ represents maximum values of $x_i, y_i, z_i$ in each segment respectively. This work resulted in 93% and 92% recalls for semantic segmentation of cracks and spalls, respectively. Also, the severe defects deeper than 7 cm achieved a recall of 98% and 99% for cracks and spalls, respectively.

In [129], the authors introduced a focal loss function and a PC registration network (PCCR-Net) network for industrial construction and infrastructure monitoring applications based on PointNet++ for segmenting components such as columns, beams, slabs, walls, concrete, and rebars. Notably, the conventional negative log-likelihood loss function of PointNet++ was replaced with the focal loss function for gradient descent. Furthermore, the authors presented a synthetic PC dataset comprising diverse precast concrete components. PCCR-Net achieved an OA of 97.47% and a mIoU of 93.12%, in comparison to PointNet++, which attained an OA of 95.17% and a mIoU of 87.68 %. However, the approach was not implemented for large PCs or PCs with geometric irregularities. [130] introduced ResPointNet++ featuring two NNs: a local aggregation operator for learning complex local structures and residual bottleneck modules to overcome gradient vanishing issues. ResPointNet++ demonstrates superior segmentation performance for indoor industrial systems, demonstrating its relevance for industrial CM and asset management use cases in comparison to PointNet++, achieving F1 scores of 98.74 % and 65.46%, respectively. The PointSift module, as proposed by [131], achieves multi-scale representation by stacking and convolving features from the nearest points across eight different spatial orientations. This versatile module can seamlessly integrate into any PointNet-based framework, enhancing the network's representation capability to achieve OA and mIoU of 88.72% and 70.23% on the S3DIS dataset. [132] defines points neighborhood in both the world space and feature space using K-means clustering and k-nearest neighbors (kNN), respectively. The learned point feature space is then structured by using pairwise distance loss and centroid loss. The mutual interaction between different points in the PC was explored by PointWeb [61] by constructing a local fully-linked web. An adaptive feature adjustment module is proposed to exchange information and refine features, then aggregate the learned features to obtain discriminative feature representation. RandLA-Net [133] introduces a lightweight NN designed to directly infer per-point semantics for large-scale PC segmentation tasks. The author incorporates a local feature aggregation module and random point sampling to retain fine-grained geometric details during object segmentation. RandLA-Net processes 200× faster than existing approaches on large PCs, surpassing SOTA semantic segmentation methods on the S3DIS and SemanticKITTI datasets. Multiple view pointwise networks (MVP-Net) [134] introduced space-filling curves and multi-rotation PC methods to expand the receptive field and efficiently aggregate the captured semantic feature. Compared to RandLA-Net [133], MVP-Net demonstrates 11 times faster performance and higher efficiency in semantic segmentation tasks using the SemanticKITTI dataset. In another study, [135] investigated the impact of neighborhood size selection for CM of bridges by segmentation of defects in 3D bridge PCs. The authors compare various sub-sampling approaches, including fast-graph, uniform, and random methods, to identify the optimal neighborhood selection strategy.

***Attention-based methods:*** These methods introduce innovative techniques for learning relations between points in PC data. [60] proposed a self-attention operator called GSA to learn relations between points. Later, the author used a task-agnostic sampling operation named GSS to replace the traditional FPS approach. This module is less sensitive to outliers, allowing a selective representative subset of points. The spatial distribution of the PC can be captured effectively using the local spatial awareness network (LSA-Net) [136]. The LSA layer hierarchically generates spatial distribution weights based on relationships in spatial regions and local structures in the PC. Based on the CRF framework proposed by [110,137] introduced an attention-based score refinement (ASR) module. This module computes weights for each point in the PC based on their initial segmentation scores, facilitating a refinement process where the scores of each point, along with those of its neighbors, are pooled together. The computed weights influence the pooling operation, offering adaptability to efficiently integrate the module into various network architectures, thereby enhancing PC segmentation. [138] used an attention-based learning module for capturing local features and semantic relations in an anisotropic manner. Subsequently, a multi-scale context-guided aggregation module was used to differentiate points in the feature space, enhancing the scene-level understanding of semantic segmentation. Global attention network (GA-Net) [139] incorporated point-independent and point-dependent GA modules for learning global contextual information across the entire PCs. Additionally, a point-adaptive aggregation block was introduced to group learned features, enhancing discriminative feature aggregation compared to linear skip connections. In [140], a semi-supervised learning (SmSL) approach called SPC-Net is introduced for segmenting various elements in tunnel PC data, including cables, segments, pipes, power tracks, supports, and tracks. A step-wise PC completion network (SPC-Net) utilizes a supervised learning model with attention mechanisms and a downsampling-up sampling structure to facilitate efficient learning and feature extraction. Furthermore, a formulated loss function is implemented to enable SPC-Net to conduct SmSL for multi-class object semantic segmentation of 3D tunnel PCs. [141] proposed a similar attention-based network called attention-enhanced sampling PC network (ASPC-Net), aimed at tunnel defect classification for infrastructure inspection and monitoring in industrial settings. ASPC-Net incorporates a weighted focal loss strategy to overcome the impact of imbalanced data, enhancing its ability to classify defects in tunnel PC datasets accurately.

***Local–global feature concatenation:*** This approach addresses the segmentation challenges posed by various object sizes and scales in large-scale industrial PCs by integrating both local and global features. Many existing methods prioritize global or local features, while hierarchical approaches often emphasize local features at the expense of global shape features. By concatenating local and global features, this approach enables comprehensive feature representation, enhancing segmentation accuracy across different object sizes and scales in large-scale industrial PC datasets. For example, PointMixer, as introduced in [142], facilitates information sharing among unstructured

3D industrial PCs by substituting token-mixing MLPs with a SoftMax function. This method aggregates features across multiple points, encompassing intra-set, inter-set, and hierarchy sets, thereby promoting effective feature fusion and information exchange within the industrial PC data. In [143], PS2-Net was introduced as a permutation-invariant approach for 3D semantic segmentation, integrating local structures and global context. The method leverages Edgeconv [8] to capture local structures and NetVLAD [144] to model global context from PCs, enabling comprehensive feature extraction for accurate segmentation of industrial components or defects. SCF-Net [145] presented a unique approach to learning spatial contextual features (SCF) tailored for large-scale industrial PCs. SCF-Net uses a local polar representation (LPR) block to construct a representation invariant to $z$-axis rotation. Neighboring representations are then aggregated via a dual-distance attentive pooling (DDAP) block to capture local features effectively. Furthermore, a global contextual feature (GCF) block utilizes local and neighborhood information to learn global context, which can be beneficial for industrial PC processing and analysis. SCF-Net's versatility allows it to integrate seamlessly into encoder–decoder architectures for 3D semantic segmentation. LGFF-Net [146] introduces a novel local feature aggregation (LFA) module to capture geometric and semantic information concurrently, preserving original data integrity during cross-augmentation. Following this, a global feature extraction (GFE) module is used to extract global features. Ultimately, local and global features are concatenated using a U-shaped segmentation structure, enhancing overall segmentation performance for industrial applications. In [147], a dual feature complementary (DFC) module is proposed to learn local features effectively. This module employs a position-aware block to move with smaller points adaptively sets, enhancing the capture of geometric features. Additionally, a global correlation mining (GCM) module is utilized to gather contextual features, further improving semantic segmentation performance. [148] highlights the integration of local feature extraction and contextual information is particularly relevant for the segmentation of overhead catenary systems in high-speed rail infrastructure, which is a critical component of industrial transportation systems. Here, local features are extracted from both the local points and their neighborhoods, followed by the aggregation of contextual information using CNN layers. Subsequently, feature enhancement and fusion techniques are applied to refine the segmentation process. Point central transformer (PointCT) [149] introduces a central-based attention mechanism and transformer architecture to address sparse annotations in PC semantic segmentation. Spatial positional encoding is introduced to focus on various geometries and scales for point representations as a valuable technique for handling the challenges associated with incomplete or limited labeling of industrial PC data. In [3], a Dempster-Shafer (D–S) evidence-based feature fusion model was employed to integrate local and global features extracted from different CNN models. The study targeted the segmentation of tunnel PC defects, encompassing cable, pipe, segment, track, and power tracks. Results showcased enhanced segmentation scores compared to raw point-based segmentation models across various baselines. This method can be used within the industrial infrastructure monitoring context. Recently, [150] introduced a transformer-based feature embedding network (3D Trans-Embed) for detecting defective industrial products for quality control and CM applications in industrial settings. The method leverages a transformer model for PC segmentation and integrates local feature embedding technology and multi-channel feature map fusion to enhance attention towards defective regions, thereby improving semantic segmentation outcomes.

**Convolution-based Methods:** These methods harness the intrinsic capabilities of CNNs to extract high-level discriminative features from complex spatial structures present in PCs. [151] used two CNNs and an RNN to conduct semantic segmentation of structural, architectural, and mechanical objects. The approach was trained and evaluated on PC data from 83 rooms representing real-world industrial and commercial buildings. In the work by [152], a novel combinational convolutional block (CCB) called PCNet++ is introduced and applied to the synthetic gear dataset (Gear-PCNet++) to detect gear defects (wear, fracture, glue, and pitting) in CM of manufacturing industries. PCNet++ replaces the convolution layer in MLP networks with the novel CCB to extract local gear information effectively while identifying its complex topology. The method outperforms PointNet, PointNet++, PointCNN, and KP-Conv on the gear PC dataset, achieving superior segmentation results. In [88], features of individual points within the PC are learned using a point-wise CNN for semantic segmentation and object recognition. Furthermore, parametric-continuous CNN (PCNN) [153] operates on non-grid data structures by employing a parameterized kernel function that spans continuous vector space. These methods utilize the spatial properties of PCs to develop point-based CNNs with spatial kernels, enabling the application of convolution operators tailored to the local structures of the PC, which could be valuable for analyzing the local structures and geometries encountered in industrial PC data. KP-Conv [85] presents a distinctive approach to 3D semantic segmentation by using radius neighborhoods as input for convolution, ensuring a consistent receptive field. This method processes these neighborhoods with weights determined spatially by a small set of kernel points. Additionally, KPConv incorporates a deformable operator to learn local shifts, enabling the customization of convolution kernels for improved alignment with the geometry of industrial PC data, potentially improving segmentation and analysis tasks. Dense connection-based kernel point network (DenseKP-NET) [154] extends the receptive field by introducing a multi-scale convolution kernel point module, facilitating the extraction of coarse-to-fine geometric features. Subsequently, a dense connection module refines these features while capturing the complex contextual information and varied object scales commonly encountered in industrial PC datasets. However, while kernel-based approaches excel in semantic segmentation, they may fail to provide ample local contextual features. To address this, [155] introduces attention kernel convolution (AKConv) to discern local contextual features while preserving object geometric shape information. In [156], a dilated point CNN (DP-CNN) is proposed to investigate the impact of the receptive field on existing point-convolution methods. DP-CNN enhances the receptive field size by aggregating features from dilated neighbors instead of KNN. Meanwhile, [157] introduces PointConvFormer, amalgamating point convolution with transformers to bolster model robustness. PointConvFormer utilizes pointwise CNN for feature extraction and computes attention weights based on feature disparities, refining convolutional weights and enhancing model performance.

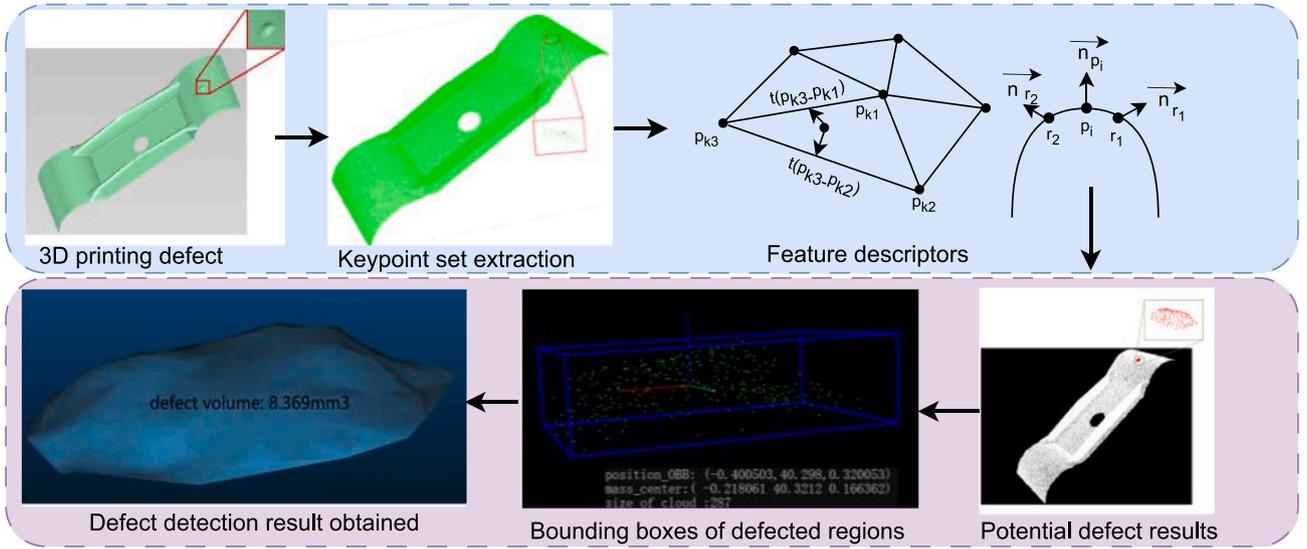

**Fig. 11.** Framework of MLS-based defect segmentation in automotive components.
*Source:* Adapted from Zhao et al. [158].

**Graph-based methods:** These methods utilize a graph as the fundamental structure for applying convolution to irregular PCs. This approach eliminates the necessity of transforming PCs into regular grids or voxels, enabling direct processing on the intrinsic graph-like nature of industrial PC data. [1] used region-growing network to segment defects, such as dents, protrusions, or scratches, on aircraft components, which is a crucial task for CM and quality control applications. The process involved smoothing the collected PC using a moving least squares (MLS) algorithm. Subsequently, curvature and normal information were collected for each point in the PC before applying the region-growing segmentation. In [158], small 3D-printed defects such as humps, collapses, and poor bridging are segmented for quality inspection using a two-step process. Initially, MLS smoothing with rough boundary removal and an improved normal rotated projection statistics (INRoPS) feature descriptor are employed to extract defect features. Subsequently, in the second stage, a neighborhood point calculation method is introduced to delineate the shape of the defects. Fig. 11 illustrates the approach employed in [158]. The approach achieved an average accuracy of 99.75% in segmenting defective regions in automotive components. A similar approach was adopted for industrial welding inspection and quality control in [159] to detect hump, pore, and bulge defects in weld bead PCs. In this study, 2D curvature analysis was utilized to detect hump defects, while 3D curvature analysis was employed to identify pore and bulge defects. A region-growing approach was then used to locate high-curvature areas for defect segmentation. Normal vectors and curvatures are essential to analyze geometrical PCs, facilitating tasks such as feature extraction, segmentation, and classification.

The local surface properties, such as normal vectors and surface curvatures for a given point $p_i = (x_i, y_i, z_i)$, can be analyzed using the eigenvalues of the covariance matrix $\boldsymbol{C}$ of the local neighborhood, defined as:

$$\boldsymbol{C} = \begin{bmatrix} p_i 1 - \bar{p} \\ \dots \\ p_{i_k} - \bar{p} \end{bmatrix}^\top \cdot \begin{bmatrix} p_i 1 - \bar{p} \\ \dots \\ p_{i_k} - \bar{p} \end{bmatrix}, i_j \epsilon N_p, \tag{14}$$

where $N_p$ is the k-nearest neighbor, $p_{i_j}$ is the $j$th neighborhood point of $p_i$ and $\bar{p}$ is centroid of the neighborhood point $N_p$. Singular value decomposition (SVD) is then performed on $\boldsymbol{C}$ to obtain eigenvalues $\lambda_m$ ($\lambda_0 < \lambda_1 < \lambda_2$). The corresponding eigenvectors $v_m(v_0, v_1, v_2)$ can be calculated as:

$$\boldsymbol{C}.v_m = \lambda_m.v_m, m\epsilon\{0, 1, 2\} \tag{15}$$

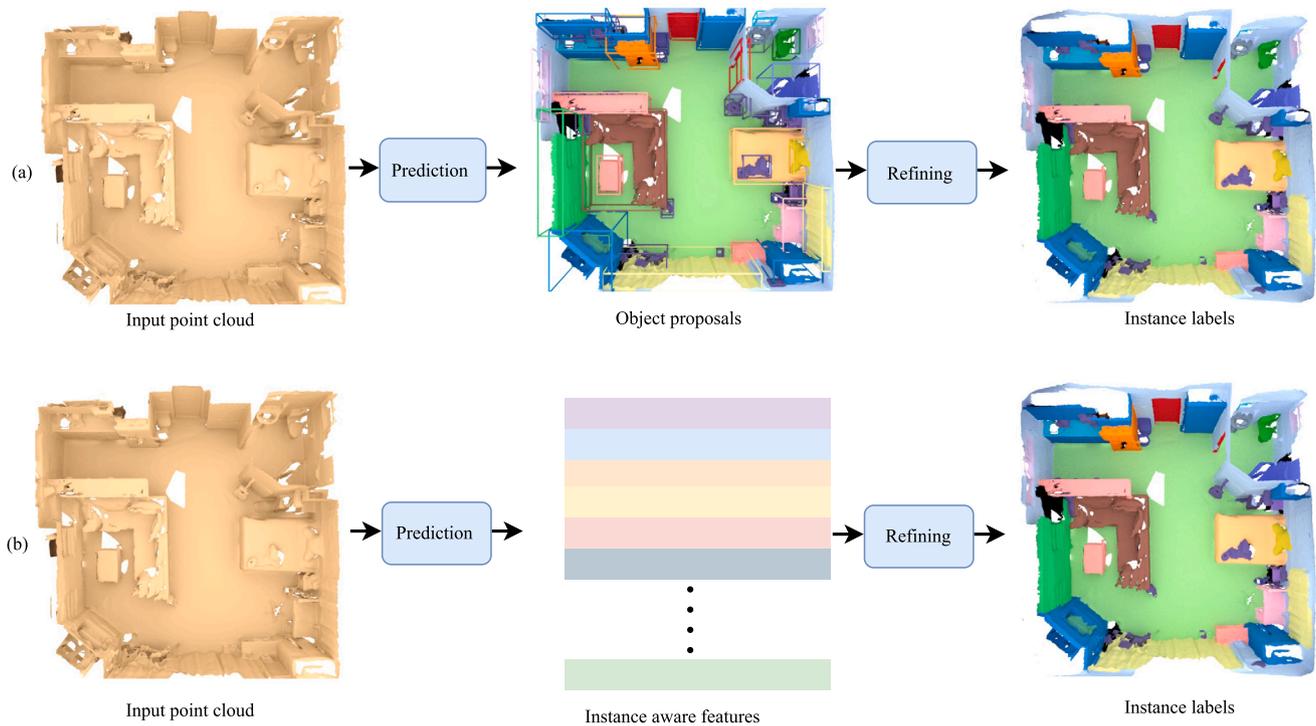

**Fig. 12.** Illustration of 3D instance segmentation frameworks on ScanNet benchmark [18]: (a) Proposal-based methods, and (b) proposal-free methods.
*Source:* Adapted from Dai et al. [18].

The normal vector $n_i$ at point $p_i$ can be estimated with the eigen vector $v_0$ corresponding to the smallest eigenvalue $\lambda_0$, given by:

$$\boldsymbol{n}_i = v_0,\tag{16}$$

Further, surface curvature $\sigma$ can be calculated using eigenvalues of the covariance matrix $\boldsymbol{C}$ given by:

$$\sigma(p_i) = \frac{\lambda_0}{\lambda_0 + \lambda_1 + \lambda_2},\tag{17}$$

These curvature values at each point in the PC are used to distinguish features and identify the damages.

DGCNN [8] treats neighboring points as a local graph and feeds it into a filter-generating network to assign edge labels. Being a transformation invariance network, DGCNN is unaffected by local point order which is well-suited for handling the irregular nature of industrial PC data. However, while it handles local points, it does not fully exploit the geometric information of neighboring points in the PC. In [160], DGCNN was examined for segmenting concrete surface defects in industrial infrastructure, such as cracks and spalls, where modifications to the loss function and data augmentation techniques, particularly flipping, were discussed to enhance performance. Later, the author [9] proposed an improved DGCNN method for semantic segmentation on concrete surfaces, effectively leveraging normal vectors and depths to detect surface defects. This normal vector-enhanced DGCNN (NVE-DGCNN) incorporates a 10-D vector by adding normal vectors $N_x$, $N_y$, and $N_z$ to the existing 7-D vectors ($x,y,z,r,g,b,d$) used in the adapted DGCNN. The method achieved recalls of 98.6% for cracks and 96.5% for spalls defects. In the study conducted by [161], the segmentation of bridge components (abutment, girder, background, pier, deck, slab, and surface) for inspection and assessment of industrial transportation infrastructure was performed using PointNet, PointCNN, and DGCNN. The results showed that DGCNN outperformed other networks, achieving an OA and mIoU of 94% and 86%, respectively. Unlike PointNet, which focuses solely on global features of input points, DGCNN incorporates information from neighboring points, enabling the generation of meaningful features to classify different bridge components based on their relationships with the surroundings. In [162] presented an enhanced version of DGCNN incorporating additional features, such as normals and colors, to segment architectural elements (arc, column, decoration, floor, door, wall, window, stairs, vault, and roof) during the inspection of buildings. The study demonstrated superior segmentation performance with custom architectural cultural heritage (ArCH) datasets with ten classes using the enhanced DGCNN and compared them to PointNet, PointNet++, PCNN, and the original DGCNN. In [10], a graph attention convolution (GAC) with learnable kernels was introduced, enabling dynamic adaptation to the structure of objects. GAC effectively learns discriminative features for semantic segmentation, offering characteristics similar to those of traditional CRF models. [163] extended the concept of GAC by introducing a cross-scale graph interaction network (CGIN) for segmenting industrial elements in remote-sensing images. CGIN used a CGI module to extract multi-scale semantic features and a boundary feature extraction (MBFE) module to learn multi-scale boundary features. Furthermore, a similarity-guided aggregation module calculates the similarity between these features, highlighting boundary information within semantic features. In [164], a simulation-to-real (sim2real) transfer learning (TL) approach is introduced, utilizing DGCNN as the backbone network for segmenting industrial elements such as pole pot, electric connection, gear container, cover, screws, magnets, armature, lower gear, and upper gear. The author also introduced a patch-based attention network to tackle imbalanced learning challenges. The network is forced to learn the same number of kernel points for all categories using an additional kernel loss. The total loss in this work is given as:

$$L_{total} = L_{seg} + \alpha L_{rot} + \beta L_{ker},\tag{18}$$

where $\beta$ is the loss weight, and $L_seg$, $L_rot$ and $L_ker$ are the segmentation, rotation and kernel losses. Here, the kernel loss is the L2 loss between the goal and the learned kernels. [165] introduced a local-attention edge convolution (LEA-Conv) layer to construct a local graph by considering neighborhood points along sixteen directions. The LAE-Conv layer assigns attention coefficients to each edge of the graph while aggregating the extracted point features through a weighted sum computation of its neighborhood. This local attention mechanism effectively captures long-range spatial contextual features, enhancing semantic segmentation's precision. [166] proposed a local–global graph CNN for semantic segmentation to capture both short and long-range dependencies within PCs. The author computes a weighted adjacency matrix for the local graph, utilizing information from neighboring points, and performs feature aggregation to capture spatial geometric features. Subsequently, these learned features are fed into a global spatial attention module to extract long-range contextual information.

### 4.2. Instance segmentation

In contrast to semantic segmentation, instance segmentation presents a more challenging task as it requires distinguishing points sharing the same semantic meaning. To address this complexity, instance segmentation methods fall into two main categories: proposal-based and proposal-free. Proposal-based instance segmentation methods can identify and segment individual industrial components, machinery parts, or defective regions within a larger PC dataset, which can then be used for comprehensive condition assessment and maintenance planning. On the other hand, proposal-free instance segmentation approaches are more suitable for handling the complex and cluttered PC data commonly encountered in industrial environments. The ability of these methods to directly segment individual instances without the need for explicit proposals could be framed as a key advantage for industrial applications, where the PC data may exhibit significant occlusions, varying object sizes, and complex spatial relationships. Fig. 12 compares proposal-based and proposal-free instance segmentation methods using the 3D ScanNet dataset [18].

#### 4.2.1. Proposal-based methods

Proposal-based instance segmentation methods can be conceptualized as a fusion of object detection and mask prediction strategies. These methods follow a top-down pipeline where the initial step involves the generation of region proposals, usually bounding boxes (BBox's), followed by predicting instance masks within these proposed regions. The pipeline encompasses multiple stages, including proposal generation, classification, and mask prediction, often integrating object detection and semantic segmentation techniques.

In [167], the authors propose Mask-Point, a multi-head region proposal extractor, to generate multiple regions of interest (ROI), allowing networks to focus on potential defective regions. However, not all 3D-ROIs belong to target defects or may overlap one another. Following this, an aggregation module comprising shared classifier, filters, and non-maximum suppression (NMS) is designed to improve the segmentation of surface defects in fiber-reinforced composites. Therefore, Mask-RCNN is used to compute the ROI probabilities through a classifier, and NMS is used to remove ROIs with lower probabilities concerning overlap and obtain final detection results. This work generated a new 3D surface defect dataset containing 120 M points and obtained mAcc and precision of 95.24% and 98.04%, respectively. [168] proposed a region-CNN (R-CNN) method by combining region proposals with features extracted from CNN to segment cracks on concrete bridges for the inspection and maintenance of industrial infrastructure, such as transportation systems and facilities. Unlike traditional CNNs based on sliding windows, R-CNN detects objects using region proposals. TL with a pre-trained model on the 50k CIFAR-10 dataset was used in the DL architecture. This pre-trained network was then fine-tuned to detect cracks on 384 collected crack images. The detected cracks were cropped and quantified using image processing techniques. Finally, the cracks were identified on an inspection map through location matching. 3D-SIS [169] is an FCN designed for 3D semantic instance segmentation using RGB-D scans. This network uses a series of CNN layers to extract 2D features for each pixel, which are then projected back onto 3D voxel grids. The RGB-D scan features are processed by 3D-CNN and aggregated into a global semantic feature map. Subsequently, 3D-Region Proposal Network (3D-RPN) and 3D-ROI layers are utilized to predict the locations of BBox's, instance masks, and object class labels. Building upon 3D-SIS, [170] applies this framework in manufacturing environments by segmenting casting defected regions (CDR) in a foundry industrial plant, introducing a non-linear topological dimension parameter to characterize the geometrical features of the segmented regions. This work evaluates cdrCNN, and DCNN defect detection networks by taking AlexNet, VGGNet-19, and ResNet-34 as backbone architectures. These networks are integrated with 3D-ROIs and instance segmentation branches to improve the AP by 10%. Generative shape proposal network (GSPN) [171] introduces a novel approach for proposal generation by reconstructing shapes from scenes, contrasting with conventional methods that regress BBox's. These generated proposals undergo refinement through a region-based PointNet, with the final labels determined by predicting point-wise binary masks for each class label, valuable for accurately segmenting and localizing individual industrial components, machinery parts, or defective regions within complex PC datasets. Importantly, GSPN incorporates a mechanism to discard trivial proposals by directly learning geometric features from the PCs. Based on PointNet++, [172] introduced 3D-BoNet, a single-stage, anchor-free, and end-to-end trainable method for achieving instance segmentation on PCs. 3D-BoNet adopts a direct regression approach to predict 3D BBox for all instances in a PC while simultaneously predicting point-level masks for each instance. The BBox prediction branch proposed in this work does not rely on pre-defined spatial anchors or RPN rather it includes 3D geometrical information along with a multi-criteria loss function. These predicted BBoxes with point and global features are fed to the point mask prediction branch to predict an accurate point-level binary mask for each instance. Gaussian instance center network (GICN) [173] utilizes Gaussian heat maps to represent the locations of instance centers distributed across the scene. By estimating the size of each instance, GICN adjusts its feature extraction process to capture relevant information within the specified neighborhood, thereby enhancing the precision and adaptability of segmentation. In [174], OccuSeg, an occupancy-aware 3D instance segmentation method, was introduced to predict point-wise instance-level segmentation. It leverages a 3D occupancy signal to predict the number of occupied pixels/voxels for each instance. This occupancy signal, learned in conjunction with feature and spatial embeddings, guides the clustering stage of 3D instance segmentation, enhancing the precision and adaptability of segmentation in industrial PC analysis, particularly for applications like asset monitoring and quality control.

**Transformer-Based Methods:** Transformer-based methods have emerged as powerful tools in various CV tasks, including semantic segmentation and instance segmentation of PCs. These methods utilize the self-attention mechanism to capture long-range semantic relationships within industrial PC data, effectively combining positional and feature information. This can be particularly valuable for industrial applications, where understanding the complex spatial and contextual relationships between different components or defects within a PC can lead to more accurate and reliable segmentation and analysis.

[175] utilizes a transformer architecture to compute object features directly from the PC data while refining predictions by updating the spatial encoding of the objects across different stages. On the other hand, segmenting objects with transformers (SOTR) [176] combines the strengths of both CNN and transformer methodologies for segmenting objects. This is achieved using a feature pyramid network (FPN) alongside twin transformers to extract lower-level features and capture long-range context dependencies for object segmentation. In the medical domain, [177] introduced a fusion of CNN and transformers termed CoTr for 3D-multi-organ segmentation. In this approach, CNNs were used for feature extraction, while a deformable transformer was utilized to capture long-range dependencies within high-resolution and multi-scale feature maps, enhancing the segmentation performance. A similar approach could be adapted for industrial inspection and CM

tasks, such as the segmentation of defects or anomalies in industrial assets. BoundaryFormer [178] used pixel-wise masks as ground truth to predict object boundaries in the form of polygons. The method evaluates the loss using an end-to-end differentiable rasterization model, enabling precise delineation of object boundaries during instance segmentation. SPFormer, as proposed by [179], is an end-to-end two-stage method designed, for instance, segmentation of PCs. Potential features extracted from the input PCs are aggregated into super points in the first stage. Subsequently, a query decoder equipped with transformers is used to directly predict instances based on these super points, facilitating efficient and accurate instance segmentation. [180] implemented improved Mask RCNN to perform instance segmentation in sewer pipelines such as breaks, deformations, and cracks. Here, the spit attention mechanism was integrated into the CNN backbone providing robust and efficient features. Also, a balanced L1 loss module was employed to improve the defect detection performance. In this work, the improved MaskRCNN was compared to Single-Shot Detector, YOLOv3, and Faster-RCNN approaches indicating improved results from the aspects of balance, loss function, and data augmentation. However, the work faces challenges in detecting smaller defects in comparison to larger defects. Mask3D [181] used stacked transformer decoders to predict instance queries, enabling the encoding of both semantic and geometric information for individual instances within a scene. While previous methods relied on instance masks for computing object queries followed by iterative refining, which often led to slow convergence, Mask3D offers an alternative approach. To alleviate the dependency on mask attention, [182] proposed a mask-attention-free transformer (MAFTr). MAFTr utilizes contextual relative position encoding for cross-attention, where position queries are iteratively updated to provide more accurate representations. Therefore, these methods precisely delineate object boundaries, efficiently predict instances, and reduce computational complexity, leading to valuable industrial PC analysis and instance segmentation tasks.

Indeed, proposal-based methods for instance segmentation offer an intuitive approach by combining object detection and mask prediction strategies. However, these methods typically involve multi-stage training processes and the need to prune redundant proposals, which can be time-consuming and computationally expensive. This complexity arises from the necessity of generating region proposals, such as BBoxs, followed by classification and mask prediction within these proposed regions. As a result, while proposal-based methods may achieve high accuracy, they often come with a significant computational cost and training overhead.

### 4.2.2. Proposal-free methods

Proposal-free methods for instance segmentation could leverage the inherent characteristics of industrial PC data, such as their spatial distribution and semantic information, to directly segment the PCs into distinct instances without relying on explicit region proposals. Instead, these methods typically use clustering techniques to group points with similar semantic meanings into distinct instances. By directly segmenting PCs into instances without the need for explicit proposals, proposal-free methods can be more computationally efficient and simpler in concept compared to proposal-based approaches. This can be framed as a more computationally efficient and conceptually simpler approach compared to the multi-stage proposal-based methods, making it potentially more suitable for real-time industrial applications.

Similarity group proposal network (SGPN) [183] is a pioneering work designed to learn features and semantic maps for individual points in a PC. This network constructs a similarity matrix that encapsulates the similarity between every pair of features within the PC. SGPN uses a double-hinge loss to enhance the discriminative features, which adjusts both the similarity matrix and the semantic segmentation results. Later, it uses a heuristic non-maximal suppression technique to merge similar points into distinct instances. However, constructing the similarity matrix demands substantial memory resources, limiting the scalability of this method. Similarly, the multi-scale affinity with sparse convolution (MASC) [184] utilizes sparse convolution to predict semantic scores for each voxel while capturing the point affinity between neighboring voxels across multiple scales. Furthermore, it uses a clustering algorithm to organize points according to the learned local similarities and the inherent mesh topology. This could be positioned as a relevant technique for segmenting complex industrial assets or infrastructure, where understanding the relationships between different components is important for condition assessment and maintenance planning. [185] proposed a structure-aware loss function to learn discriminative embeddings for each instance by considering the similarity between geometric and embedding information. The author proposed attention-based kNN to refine the learned features by grouping information from neighbors while eliminating the quantization error caused by the 3D voxel.

Several methods have been proposed, such as integrating semantic category and instance label prediction into a single task. Milestones in 3D PC instance segmentation, including both proposal-based and proposal-free methods, are depicted in Fig. 13. [186] integrates the advantages of both instance and semantic segmentation through an end-to-end learnable module called associatively segmenting instances and semantics (ASIS). The ASIS module incorporates semantic-aware point-level embedding to achieve instance segmentation and performs instance fusion to obtain semantic segmentation simultaneously. [187] introduced a joint instance and semantic segmentation (JISS) module, which combines instance and semantic segmentation to generate discriminative features. To address the large memory consumption of JSNet, the authors proposed dynamic filters for convolution (DF-Conv) on PCs. Based on JSNet, DFConv, and an enhanced JISS (JISS*) module, [188] introduced JSNet++ to enhance instance segmentation. These methods emphasize the impact of joint learning for industrial applications, where understanding both the semantic context and the individual instances of components or defects can provide a more comprehensive understanding of the overall system or asset condition. 3D-Multi proposal aggregation (3D-MPA) [189] presents a technique for predicting object proposals using semantic features derived from a sparse volumetric backbone network. In contrast to conventional non-maximum suppression (NMS), this method employs the MPA strategy, based on learned features, to derive semantic instances from the generated object proposals. [71] identified structural defects in 3D PC of concrete bridges by detecting, mapping, and extracting defects through instance clustering using a CNN-based detection method called DetectionHMA. In the detection stage, semantic segmentation is performed on all images to return class probabilities, which are then mapped from 2D to a dense cloud, yielding segmented PCs. Finally, the segmented PC is clustered, and the respective sub-clouds are transformed into defect instances. Three CNN-based approaches – TopoCrack [190], nnU-Net [191], and DetectionHMA – were compared. DetectionHMA showed better performance for cracks with a mIoU of more than 90%, while nnU-Net performed well for areal anomalies such as spalls and corrosion. However, instance segmentation, measured in AP, was distinctly low, indicating a need for more advanced quantitative analyses. Nevertheless, they leverage semantic features and clustering to derive instance segmentation demonstrating relevant techniques for the detection and mapping of structural defects in industrial infrastructure, such as bridges or concrete surfaces, contributing to improved CM and maintenance-planning.

**Grouping-Based Methods:** In contrast to proposal-based methods, grouping-based methods follow a bottom-up pipeline approach, leveraging the inherent characteristics of industrial PC data, such as semantic labels and instance center offsets, to directly group points into distinct instances without the need for explicit region proposals [192, 193]. This can be framed as a computationally efficient approach that aligns well with the requirements of industrial applications, where real-time processing and analysis of PC data are often crucial.

[194] proposed a multi-task segmentation algorithm (MSA) to learn unique feature embeddings for each instance by leveraging grouping

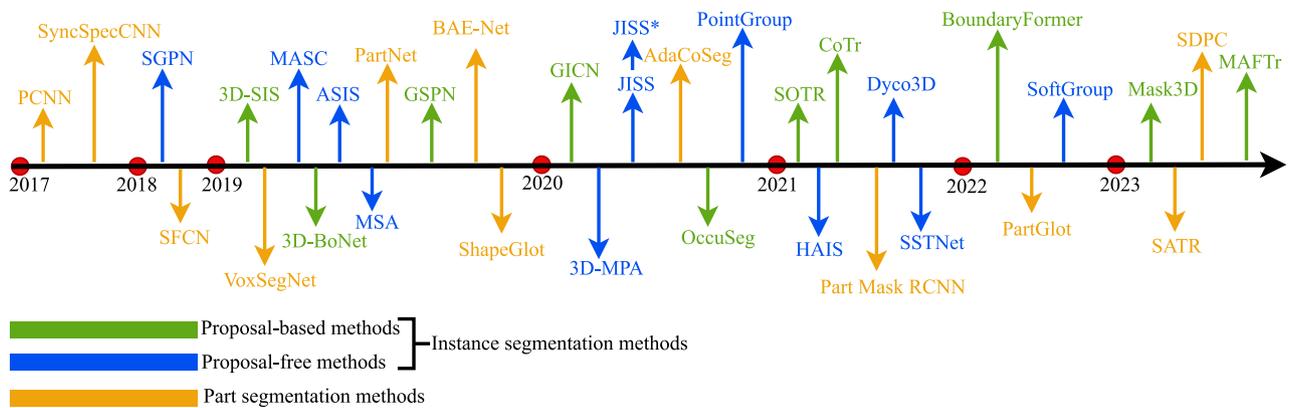

**Fig. 13.** Chronological overview of the most relevant DL-based 3D instance and part segmentation methods.

or clustering information associated with individual objects. Point-Group [195] focuses on grouping points by identifying the void space between distinct objects. The authors proposed a two-branch network capable of extracting point features, predicting semantic labels, and computing offsets concurrently to achieve this. These offsets are then employed to relocate each point towards its corresponding instance centroid. Based on PointGroup, [196] introduced a clustering-based framework called hierarchical aggregation for 3D IS (HAIS) to produce detailed instance predictions while effectively filtering out noisy points within instance predictions. PointGroup and its subsequent extensions could be positioned as relevant techniques for industrial PC analysis. The grouping of points by identifying the void space between distinct objects, as well as the effective filtering of noisy points, can contribute to more accurate and robust instance segmentation for asset management and quality control applications. Dyco3D [197] introduced dynamic convolution kernels, which encode category-specific context by utilizing a sub-network to explore homogeneous points showing close votes, for instance, centroids and sharing the semantic labels. The parallel decoding of instance masks is accomplished by convolving the generated class-specific filters with coordinate information. SST-Net [192] introduced a semantic super-point tree, where each super-point represents a geometrically homogeneous neighborhood. This method utilizes tree traversal for object proposal by splitting non-similar nodes in this semantic super-point tree. SoftGroup [193] addresses errors arising from hard semantic predictions by performing grouping based on semantic scores. The method uses a top-down refinement module using U-Net to improve positive samples while suppressing false positives introduced by incorrect semantic predictions. This top-down refinement can contribute to more accurate and reliable instance segmentation, ultimately benefiting industrial systems in CM tasks.

In summary, while proposal-free methods alleviate the computational burden associated with region-proposal mechanisms, they often exhibit lower objectness in the resulting grouped instance segments. This limitation stems from their inherent inability to explicitly detect object boundaries, leading to less precise delineation of individual objects within the PC.

### 4.3. Part segmentation

Part segmentation involves categorizing the PC into distinct groups, each representing a specific physical part of the object. It is crucial for industrial systems, where accurately categorizing the distinct physical parts of an object or asset can provide valuable insights for maintenance, inspection, and optimization. However, part segmentation encounters two significant challenges. First, parts with the same semantic label may exhibit considerable geometric variation and ambiguity. Second, objects with identical semantic meanings may consist of different numbers of constituent parts. Several milestone 3D PC part segmentation methods have been illustrated in Fig. 13.

VoxSegNet [201] introduced a spatial pose extraction (SDE) module to extract multi-scale discriminative features from sparse volumetric data. These learned features are contextually selected and aggregated through an attention feature aggregation (AFA) module, ensuring dense prediction with semantic consistency and enhanced accuracy. It can contribute to more accurate and consistent part segmentation for complex industrial assets, enabling better understanding of their internal structure and condition. PartNet [202] introduces a top-down, fine-grained, and hierarchical approach to part segmentation. Unlike conventional methods that segment shapes into a fixed set of labels, PartNet formulates part segmentation as a cascade binary labeling process. This methodology decomposes the input PC into an arbitrary number of parts determined by the underlying geometric structures, essential for maintenance planning, quality control, and optimization of manufacturing processes. [203] introduced an end-to-end network called projective CNNs (PCNNs), which combines FCNs and surface-based CRFs to achieve part segmentation of 3D shapes. The authors selected images from multiple views to ensure optimal surface coverage and fed them into the network to generate per-part confidence maps. These confidence maps are then aggregated using surface-based CRFs to label the entire shape. However, dealing with different shapes resulted in different nearest-neighbor graphs in the PC, posing challenges for weight sharing among convolution kernels across various shapes. To address this challenge, synchronized spectral CNN (Sync-SpecCNN) [204] uses a spectral network for convolution, allowing weight sharing across different non-isometric shapes. The multi-view information and the ability to handle non-isometric shapes can contribute to more robust and accurate part segmentation of industrial PC data, where the geometry and structure of assets or components can vary significantly. Additionally, [205] introduced part segmentation on 3D meshes using shape FCNs (SFCNs). The author utilized SFCNs to process low-level geometric features and refined the segmentation outcomes through feature voting-based multi-level graph cuts. In [206], the authors proposed Part-Mask RCNN for predicting shape categories, BBoxes, object masks, and object part masks in RGB-D images. The authors utilized a voting-based pose estimation algorithm on semantic information of the objects to obtain part segmentation. This combination of shape category prediction, BBox estimation, object mask generation, and part mask segmentation can provide a comprehensive understanding of the structure and composition of industrial assets or components. Later, [207] proposed an adaptive shape co-segmentation (AdaCoSeg) network to address the challenges associated with retraining and adapting to newer input datasets which can be used for handling the evolving nature of industrial assets and the variability in PC data collected from different sources. AdaCoSeg takes a set of unsegmented PC shapes as input and iteratively minimizes the group

**Table 3**
Summary of PC-based defect segmentation in industrial systems.

| Ref. | Application | Classes | Method | Results | Points/Objects |
|---|---|---|---|---|---|
| [162] | Semantic segmentation of architectural elements | 10 classes: column, decoration, door, arc, wall, window, stairs, vault, and roof | PointNet | OA = 21.6, mIoU = 10.9 | 114 M |
| | | | PointNet++ | OA = 24.5, mIoU = 18.0 | |
| | | | DGCNN | OA = 54.7, mIoU = 35.8 | |
| | | | PCNN | OA = 39.5, mIoU = 33.1 | |
| | | | Modified DGCNN | OA = 71.6, mIoU = 37.7 | |
| [161] | Semantic segmentation of architectural elements | 6 classes: abutment, slab, pier, girder, surface, and background | PointNet | OA = 93.8, mIoU = 84.3 | N/A |
| | | | PointCNN | OA = 92.6, mIoU = 76.8 | |
| | | | DGCNN | OA = 94.5, mIoU = 86.9 | |
| [168] | Segmentation of bridge elements | 3 classes: deck, pier, and background | PointNet | OA = 94, mIoU = 84 | 50,000 |
| [127] | Semantic segmentation of underwater pipe | 3 classes: pipe, valve, and background | PointNet | F1 score = 89.3 | 262 |
| [160] | Segmentation of concrete surface | 3 classes: crack, spall, and normal | DGCNN | OA = 98, F1 score = 98 | 49 M |
| [117] | Segmentation of synthetic concrete defects | 2 classes: cracks, and spalls | InspectionNet | mAcc = 96.46 | 12,000 |
| [100] | Segmentation of bridge elements | 2 classes: slab and pier | DGCNN | OA = 95.9, mIoU = 71.1 | 447 M |
| | | | PointNet | OA = 84.4, mIoU = 45.9 | |
| [128] | Semantic segmentation of concrete bridge elements | 3 classes: cracks, spalls, and normal | SNEPointNet++ | OA = 95.9, mIoU = 83.26 | 27 M |
| | | | Adaptive PointNet++ | OA = 97.12, mIoU = 63.36 | |
| [152] | Segmentation of gear | 5 classes: basic gear, fracture, glue, wear, pitting | Gear-PCNet++ | OA = 99.53, mIoU = 98.97 | 10,000 |
| | | | PointNet++ | OA = 99.29, mIoU = 98.50 | |
| | | | PointCNN | OA = 99.43, mIoU = 98.76 | |
| | | | KPConv | OA = 99.64, mIoU = 97.50 | |
| [129] | Semantic segmentation in precast concrete rebar | 4 classes: column, beam, slab, and wall | PCCR-Net | OA = 97.47, mIoU = 93.12 | 342 |
| | | | PointNet++ | OA = 95.17, mIoU = 87.68 | |
| [198] | Panoptic segmentation in railway infrastructure | 7 classes: informative signs, masts, traffic lights, traffic signs, cables, droppers and rails | PointNet++ | OA = 95.34, mIoU = 80.3 | 4.5 M |
| [140] | Semi-supervised segmentation of 3D tunnel elements | 6 classes: cable, segment, pipe, power track, support, track | SPCNet | OA = 97.23, mIoU = 97.41 | 32 M |
| [141] | Segmentation of 3D tunnel elements | 7 classes: cable, segment, pipe, power track, seepage, support, track | ASPCNet | OA = 97.58, mIoU = 89.80 | 34 M |
| [199] | Segmentation of 3D tunnel elements | 7 classes: cable, segment, pipe, power track, seepage, support, track | DGCNN | F1 = 91.9, mIoU = 97.5 | 34 M |
| | | | PointNet | F1 = 98.1, mIoU = 96.3 | |
| [148] | Segmentation of overhead catenary systems (OCS's) in high-speed rails | 8 classes: cantilevers, catenary wires, contact wires, droppers, insulators, poles, registration arms, steady arms | KNN+CNN | Precision = 97.50, mIoU = 94.84 | 16 M |
| | | | PointNext | Precision = 96.49, mIoU = 93.39 | |
| | | | PointNet++ (SSG) | Precision = 96.42, mIoU = 93.06 | |
| | | | PointNet | Precision = 94.52, mIoU = 89.18 | |
| [164] | Segmentation of industrial elements | 9 classes: pole pot, electric connection, gear container, cover, screws, magnets, armature, lower gear, upper gear | DGCNN | mIoU (real) = 93.75, mIoU (Simulation) = 98.01 | 5.2 M |
| [130] | Industrial indoor LiDAR dataset | 6 classes: I-beam, pipe, pump, rectangular beam, and tank | PointNet | OA = 53.0, mIoU = 21.1 | 5 M |
| | | | PointNet++ | OA = 70.6, mIoU = 45.5 | |
| | | | ResPointNet++ | OA = 94.0, mIoU = 87.3 | |
| [150] | Segmentation of vegetation and industrial products | 6 classes: potatoes, carrots, peaches, cookies, bagels, cable, 6 industrial products: cable gland, dowel, tyres, foam, and ropes | 3D Trans-Embed | F1-score = 83.32, Precision = 87.82 | 4000 |
| | | | PCT | F1-score = 72.71, Precision = 73.48 | |
| | | | PointNet++ | F1-score = 65.15, Precision = 69.18 | |
| [200] | Instance segmentation of different object shapes in an oil refinery, petrochemical plant, and warehouse | 8 classes: cylinders, angles, channels, I-beams, elbows, flanges, valves and miscellaneous | CLOI-NET | mPrec = 73.2, mRec = 71.1 | N/A |
| | | | ASIS | mPrec = 74, mRec = 24.9 | |
| [151] | Semantic segmentation of structural, architectural, and mechanical objects | 6 classes: beam, ceiling, column, floor, pipe, and wall | CNN+RNN | mAcc = 86.13 | 10.8 M |
| | | | CNN | mAcc = 84.70 | |
| [167] | Segmentation of surface defects in fiber composites | 2 classes: void defects and surface defects | MaskPoint | mAcc = 99.97, mIoU = 94.02 | 120 M |
| | | | PointNet++ | mAcc = 99.50, mIoU = 58.81 | |
| | | | PointNet | mAcc = 99.41, mIoU = 62.72 | |
| | | | KPConv | mAcc = 99.32, mIoU = 49.53 | |
| | | | PointTransformer | mAcc = 99.38, mIoU = 49.83 | |

**Table 3** (*continued*).

| Ref. | Application | Classes | Method | Results | Points/Objects |
|------|-------------|---------|--------|---------|----------------|
| [66] | Defect segmentation in concrete sewer pipes | 5 classes: 3 circular defects of varying diameter, square and triangular defect | Improved PointNet++ | mIoU = 94.15 | 1.4 M |
| | | | PointNet++ | mIoU = 82.69 | |
| | | | Point Transformer | mIoU = 86.31 | |
| [9] | Semantic segmentation in concrete surfaces | 3 classes: cracks, spalls, and normal | NVE-DGCNN | mPrec = 96.99, Recall = 98.11, mIoU = 95.24 | 49 M |
| | | | SNEPointNet++ | mPrec = 87.5, Recall = 94.6, mIoU = 83.26 | |
| [109] | Defect segmentation in railway tracks | 3 classes: scratches, peeling, and cracks | T-PCIF | mAcc = 86.27, mIoU = 70.18 | 618 |
| [159] | Defect segmentation in wire and arc AM | 3 classes: pores, bulges, and humps | Region-growing | mAcc = 90.5, mPrec = 92, mRec = 90.4 | 899 |
| [180] | Instance segmentation in sewer pipes inspection | 3 classes: breaks, cracks, and deformations | Improved Mask RCNN | mPrec = 92.5 mIoU = 92.7 | 1744 |
| [71] | Instance segmentation in concrete bridges | 4 classes: spalls, cracks, corrosion, and background | DetectionHMA | mIoU = 99.0. | 108.4 M |
| | | | nnU-Net | mIoU = 90.6 | |
| | | | TopoCrack | mIoU = 85.6 | |

consistency loss to produce shape part labels. Unlike traditional CFR methods, the authors refine and denoise the part proposals using a pre-trained part-refinement network. The branched auto-encoder network (BAE-NET) [208] tackles the 3D shape co-segmentation task by framing it as a representation learning challenge. This approach aims to discover the most concise part representations by minimizing the shape reconstruction loss. Each network branch is dedicated to learning a condensed representation for a particular part shape using an encoder–decoder architecture. The features acquired from each branch, combined with the point coordinates, are fed into the decoder to produce a binary value indicating whether the point belongs to that part. In the medical domain, [209] proposed a shape-aware segmentation (SAS) technique for processing MRI imaging scans. This method imposes geometric constraints on both labeled and unlabeled input data. It involves learning a shape-aware representation using a signed distance map (SDM) approach. Following this, the obtained predictions undergo refinement through an adversarial loss. Similar approaches leveraging geometric constraints and adversarial learning can be adapted for part segmentation of industrial PC data, where maintaining shape consistency and refining the segmentation results are crucial for accurate asset analysis and condition assessment. Based on the ShapeGlot framework [210], PartGlot [211] utilizes a transformer-based attention mechanism to understand the regions corresponding to semantic parts by leveraging linguistic descriptions. [212] proposed a soft density peak clustering (SDPC) algorithm tailored for 3D shape segmentation. [213] developed a segmentation assignment with topological re-weighting (SATR) to achieve part segmentation from the predicted multi-view BBox's. Firstly, Gaussian geodesic re-weighting is performed to adjust weights by considering the geodesic distance from potential segment centers. Secondly, a graph kernel is used to refine the inferred weights considering the neighbor's visibility. These two techniques are combined to achieve state-of-the-art 3D shape segmentation for fine-grain queries. It could be framed as advanced techniques that can contribute to more accurate and robust part segmentation of industrial PCs, particularly for complex assets or components with intricate geometric structures. [213] used geodesic curves for discriminative modeling of the object shapes within an NN framework. The method involves selecting pairs of 3D points on depth images to compute surface geodesics. The approach leverages a large training set of geodesics created using minimal ground truth instance annotations, where each geodesic is labeled binary to indicate whether it belongs entirely to one instance segment. An NN is then trained to classify geodesics based on these labels. During inference, geodesics are generated from selected seed points in the test depth image, and a convex hull is constructed for points classified by the neural network (NN) as belonging to the same instance, thereby achieving instance segmentation.

In summary, 3D part segmentation allows a fine-grained understanding of objects by categorizing the PCs into distinct parts, providing geometric information about the objects in the scene. Also, it contributes to semantic interpretation, enabling systems to recognize and label structural components of objects. However, objects from the same semantic class may have significant intra-class variability, making it challenging to define consistent part boundaries. Table 3 present the outcomes of defect segmentation for industrial systems.

### 4.4. Summary

This section presents key challenges and research directions in processing 3D PCs for defect classification and segmentation:

- **Data Collection**: Collecting extensive datasets for industrial applications is inherently challenging due to the need for varied environments, equipment, and scenarios. Achieving a representative dataset requires capturing data from different times, locations, and conditions to ensure the model generalizes well to real-world situations. To address this, creating synthetic datasets through simulation environments can supplement real-world data collection.
- **Labeling**: Annotation and ground truth labeling, especially for complex industrial defects, are labor-intensive and prone to errors. Domain adaptation and TL can overcome the need for extensive labeled datasets, especially in scenarios with limited labeled data availability. Techniques like generative adversarial networks (GANs) or other generative models can augment datasets, expand training data and improve model generalization.
- **Noisy and Incomplete Data**: The quality of collected PC data, particularly from outdoor scenes, is often compromised by various factors, including noise, outliers, and missing points. These issues pose significant challenges for accurate data analysis and interpretation. Efforts are being made to address these challenges by developing advanced methods, such as generative approaches that can synthesize realistic data to fill in gaps and reduce noise, and discriminative methods that enhance the ability to distinguish between true data points and outliers.
- **Scalability**: High-resolution PCs can encompass several million points, leading to significant challenges in data processing, storage, and analysis. To address these issues, methods such as downsampling reduce the number of points while preserving essential geometric features. Additionally, spatial partitioning techniques such as octrees and voxel grids organize data into hierarchical structures for efficient processing. However, it is still necessary to continually develop new strategies to manage PC sizes effectively, ensuring efficient handling and processing of large-scale datasets.
- **Computational Efficiency**: Efficient processing of PC data requires reducing computational load and optimizing algorithms to handle the high-dimensional, and unstructured PC data. Development of lightweight DL models such as PointNet and PointNet++

has helped to minimize this by reducing the number of parameters and computation complexity. Additionally, utilizing GPUs for parallel processing and distributed computing frameworks can significantly accelerate these processing methods, enabling faster and more efficient analysis of PC data.

- **Information Loss**: The projection-based methods often adopt network architectures similar to their 2D image counterparts. However, a key limitation of these methods is the loss of information due to the conversion from 3D to 2D projection. On the other hand, volumetric-based representations encounter challenges with significantly increased computational and memory costs attributed to the cubic growth in resolution. Addressing these issues, sparse convolution methods leveraging indexing structures emerge as a promising solution that needs further exploration. Popular for defect classification and segmentation, point-based networks lack explicit, neighboring information, relying on costly neighbor-searching mechanisms, thus limiting their efficiency.
- **Imbalanced Data**: Learning from imbalanced data remains challenging, with approaches struggling in minority classes despite strong overall performance. Novel techniques are needed to handle imbalanced datasets effectively, such as data augmentation, class balancing, or specialized loss functions (weighted cross-entropy loss and focal loss).
- **Insufficient Data**: The literature presents numerous studies dedicated to defect segmentation within general objects or spaces using 3D PC data. However, a significant gap exists in the research concerning detecting damages within industrial systems using 3D PC data. Despite the promising outcomes of semantic segmentation in PC analysis for damage detection in industrial systems, its effectiveness heavily relies on the availability of comprehensive datasets for model training. Unfortunately, the literature lacks sufficient datasets for defect estimation in industrial systems, underscoring the urgent need to collect abundant and efficient data for this purpose.

While considerable research has been done in PC shape classification and object segmentation across fields like robotics, autonomous vehicles, and other CV applications using 3D PC data, the detection of damages in industrial systems remains relatively underexplored. This gap signifies a significant opportunity for future research to devise specialized methods and models specifically addressing the distinct challenges of CM in industrial systems.

## 5. Conclusion

This paper provides a comprehensive survey and discussion of DL-based PC classification and segmentation, with a specific focus on their applications in industrial systems. The review outlines the significance of PC data and the unique challenges associated with processing this unstructured information. The paper presents a detailed taxonomy of the existing DL methods for processing 3D PC data, categorizing them into view-based, volumetric-based, and direct point-based approaches. Direct PC-based methods process the original 3D PC data, leveraging the rich information in the PC representation. These methods can overcome the potential information loss associated with the projection and discretization steps required by multi-view and volumetric-based approaches. Therefore, direct PC-based methods can be considered a promising future research direction, as they can better preserve the inherent geometric properties and spatial relationships within the 3D data. Moreover, recent advancements in transformer-based architectures such as MVCVT, PCT, and BERT have shown great potential for PC tasks.

The paper also delves into various DL-based methods for 3D PC segmentation, including semantic segmentation, instance segmentation, and part segmentation. Instance segmentation represents a challenging task in CV, combining target detection and semantic segmentation. While limited studies focus on 3D instance segmentation of defects in industrial systems, the future holds promising prospects for DL models in this domain. The paper compares the performance of the existing methods for PC shape classification and segmentation, providing insights into their strengths and limitations for industrial systems.

While the diversity of real-world scenes, including indoor environments, roads, railways, and buildings, presents opportunities for PC classification and segmentation, it also poses challenges in determining the specific advantages of the numerous methods. Researchers need to carefully select classification algorithms that align with the unique requirements of a given industrial scenario, emphasizing the need for adaptability to diverse real-world conditions. This challenge is further exacerbated by the current scarcity of suitable datasets, highlighting the importance of expanding and diversifying training data to comprehensively evaluate and improve the efficacy of PC processing techniques across different industrial applications.

In the future, researchers should focus on developing more robust and generalizable DL models that can handle the complex challenges encountered in real-world industrial environments, such as noisy data, occlusions, and varying operating conditions. In parallel, advancing computational efficiency and scalability of PC processing techniques is crucial to enable their deployment in industrial settings. This includes developing more lightweight NN architectures, such as PointNet and PointNet++-based architectures, and leveraging efficient convolution operations. Furthermore, exploring TL and domain adaptation strategies can enhance the adaptability of DL models across diverse industrial assets and defect types, particularly in case of data scarcity. By leveraging knowledge gained from related domains or simulated data, researchers can improve the generalization capabilities of these models, enabling their application in a wider range of industrial scenarios. By exploring these research directions, researchers can ultimately lead to more robust, adaptive, and practical solutions for defect detection and classification in industrial systems.

## CRediT authorship contribution statement

**Anju Rani:** Writing – review & editing, Writing – original draft, Visualization, Software, Methodology, Investigation, Formal analysis, Conceptualization. **Daniel Ortiz-Arroyo:** Writing – review & editing, Supervision, Project administration, Funding acquisition, Conceptualization. **Petar Durdevic:** Writing – review & editing, Validation, Supervision, Resources, Project administration, Investigation, Funding acquisition, Conceptualization.

## Declaration of competing interest

The authors declare that they have no known competing financial interests or personal relationships that could have appeared to influence the work reported in this paper.

## Data availability

No data was used for the research described in the article.

## Acknowledgments

This research was supported by Aalborg University, Liftra ApS (Liftra), and Dynamica Ropes ApS (Dynamica) in Denmark under the Energiteknologiske Udviklings- og Demonstrationsprogram (EUDP) program through project grant number 64021-2048.

## References

[1] Igor Jovančević, Huy-Hieu Pham, Jean-José Orteu, Rémi Gilblas, Jacques Harvent, Xavier Maurice, Ludovic Brèthes, 3D point cloud analysis for detection and characterization of defects on airplane exterior surface, J. Nondestruct. Eval. 36 (2017) 1–17.

[2] Yulan Guo, Hanyun Wang, Qingyong Hu, Hao Liu, Li Liu, Mohammed Bennamoun, Deep learning for 3d point clouds: A survey, IEEE Trans. Pattern Anal. Mach. Intell. 43 (12) (2020) 4338–4364.

[3] Huang Zhang, Changshuo Wang, Shengwei Tian, Baoli Lu, Liping Zhang, Xin Ning, Xiao Bai, Deep learning-based 3D point cloud classification: A systematic survey and outlook, Displays (2023) 102456.

[4] Qian Wang, Min-Koo Kim, Applications of 3D point cloud data in the construction industry: A fifteen-year review from 2004 to 2018, Adv. Eng. Inform. 39 (2019) 306–319.

[5] Yifan Feng, Zizhao Zhang, Xibin Zhao, Rongrong Ji, Yue Gao, Gvcnn: Group-view convolutional neural networks for 3d shape recognition, in: Proceedings of the IEEE Conference on Computer Vision and Pattern Recognition, 2018, pp. 264–272.

[6] Peng-Shuai Wang, Yang Liu, Yu-Xiao Guo, Chun-Yu Sun, Xin Tong, O-cnn: Octree-based convolutional neural networks for 3d shape analysis, ACM Trans. Graph. 36 (4) (2017) 1–11.

[7] Shaoshuai Shi, Xiaogang Wang, Hongsheng Li, Pointrcnn: 3d object proposal generation and detection from point cloud, in: Proceedings of the IEEE/CVF Conference on Computer Vision and Pattern Recognition, 2019, pp. 770–779.

[8] Yue Wang, Yongbin Sun, Ziwei Liu, Sanjay E. Sarma, Michael M. Bronstein, Justin M. Solomon, Dynamic graph cnn for learning on point clouds, ACM Trans. Graph. (tog) 38 (5) (2019) 1–12.

[9] Fardin Bahreini, Amin Hammad, Dynamic semantic segmentation of concrete defects and as-inspected modeling, Autom. Constr. 159 (2024) 105282.

[10] Lei Wang, Yuchun Huang, Yaolin Hou, Shenman Zhang, Jie Shan, Graph attention convolution for point cloud semantic segmentation, in: Proceedings of the IEEE/CVF Conference on Computer Vision and Pattern Recognition, 2019, pp. 10296–10305.

[11] Shuo Chen, Tan Yu, Ping Li, Mvt: Multi-view vision transformer for 3d object recognition, 2021, arXiv preprint arXiv:2110.13083.

[12] Jie Li, Zhao Liu, Li Li, Junqin Lin, Jian Yao, Jingmin Tu, Multi-view convolutional vision transformer for 3D object recognition, J. Vis. Commun. Image Represent. 95 (2023) 103906.

[13] Jianhui Yu, Chaoyi Zhang, Heng Wang, Dingxin Zhang, Yang Song, Tiange Xiang, Dongnan Liu, Weidong Cai, 3D medical point transformer: Introducing convolution to attention networks for medical point cloud analysis, 2021, arXiv preprint arXiv:2112.04863.

[14] Daniel Munoz, J. Andrew Bagnell, Nicolas Vandapel, Martial Hebert, Contextual classification with functional max-margin markov networks, in: 2009 IEEE Conference on Computer Vision and Pattern Recognition, IEEE, 2009, pp. 975–982.

[15] Franz Rottensteiner, Gunho Sohn, Jaewook Jung, Markus Gerke, Caroline Baillard, Sebastien Benitez, Uwe Breitkopf, The ISPRS benchmark on urban object classification and 3D building reconstruction, ISPRS Ann. Photogramm. Remote Sens. Spat. Inf. Sci. 1 (1) (2012) 293–298, I-3.

[16] Andrés Serna, Beatriz Marcotegui, François Goulette, Jean-Emmanuel Deschaud, Paris-rue-Madame database: a 3D mobile laser scanner dataset for benchmarking urban detection, segmentation and classification methods, in: 4th International Conference on Pattern Recognition, Applications and Methods ICPRAM 2014, 2014.

[17] Bruno Vallet, Mathieu Brédif, Andrés Serna, Beatriz Marcotegui, Nicolas Paparoditis, TerraMobilita/iQmulus urban point cloud analysis benchmark, Comput. Graph. 49 (2015) 126–133.

[18] Angela Dai, Angel X. Chang, Manolis Savva, Maciej Halber, Thomas Funkhouser, Matthias Nießner, ScanNet: Richly-annotated 3D reconstructions of indoor scenes, in: Proc. Computer Vision and Pattern Recognition, CVPR, IEEE, 2017.

[19] Iro Armeni, Ozan Sener, Amir R. Zamir, Helen Jiang, Ioannis Brilakis, Martin Fischer, Silvio Savarese, 3D semantic parsing of large-scale indoor spaces, in: Proceedings of the IEEE Conference on Computer Vision and Pattern Recognition, 2016, pp. 1534–1543.

[20] Timo Hackel, Nikolay Savinov, Lubor Ladicky, Jan D. Wegner, Konrad Schindler, Marc Pollefeys, Semantic3d. net: A new large-scale point cloud classification benchmark, 2017, arXiv preprint arXiv:1704.03847.

[21] Xavier Roynard, Jean-Emmanuel Deschaud, François Goulette, Paris-Lille-3D: A large and high-quality ground-truth urban point cloud dataset for automatic segmentation and classification, Int. J. Robot. Res. 37 (6) (2018) 545–557.

[22] A. Geiger, P. Lenz, R. Urtasun, Are we ready for autonomous driving? The KITTI vision benchmark suite, in: Proc. of the IEEE Conf. on Computer Vision and Pattern Recognition, CVPR, 2012, pp. 3354–3361.

[23] Weikai Tan, Nannan Qin, Lingfei Ma, Ying Li, Jing Du, Guorong Cai, Ke Yang, Jonathan Li, Toronto-3D: A large-scale mobile lidar dataset for semantic segmentation of urban roadways, in: Proceedings of the IEEE/CVF Conference on Computer Vision and Pattern Recognition Workshops, 2020, pp. 202–203.

[24] Nina Varney, Vijayan K. Asari, Quinn Graehling, DALES: A large-scale aerial LiDAR data set for semantic segmentation, in: Proceedings of the IEEE/CVF Conference on Computer Vision and Pattern Recognition Workshops, 2020, pp. 186–187.

[25] Holger Caesar, Varun Bankiti, Alex H. Lang, Sourabh Vora, Venice Erin Liong, Qiang Xu, Anush Krishnan, Yu Pan, Giancarlo Baldan, Oscar Beijbom, nuScenes: A multimodal dataset for autonomous driving, 2020.

[26] Zhirong Wu, Shuran Song, Aditya Khosla, Fisher Yu, Linguang Zhang, Xiaoou Tang, Jianxiong Xiao, 3D shapenets: A deep representation for volumetric shapes, in: Proceedings of the IEEE Conference on Computer Vision and Pattern Recognition, 2015, pp. 1912–1920.

[27] Angel X. Chang, Thomas Funkhouser, Leonidas Guibas, Pat Hanrahan, Qixing Huang, Zimo Li, Silvio Savarese, Manolis Savva, Shuran Song, Hao Su, et al., Shapenet: An information-rich 3d model repository, 2015, arXiv preprint arXiv:1512.03012.

[28] Jiachen Sun, Qingzhao Zhang, Bhavya Kailkhura, Zhiding Yu, Chaowei Xiao, Z. Morley Mao, Benchmarking robustness of 3D point cloud recognition against common corruptions, 2022, arXiv preprint arXiv:2201.12296.

[29] Mikaela Angelina Uy, Quang-Hieu Pham, Binh-Son Hua, Duc Thanh Nguyen, Sai-Kit Yeung, Revisiting point cloud classification: A new benchmark dataset and classification model on real-world data, in: International Conference on Computer Vision, ICCV, 2019.

[30] Meida Chen, Qingyong Hu, Zifan Yu, Hugues Thomas, Andrew Feng, Yu Hou, Kyle McCullough, Fengbo Ren, Lucio Soibelman, STPLS3D: A large-scale synthetic and real aerial photogrammetry 3D point cloud dataset, in: 33rd British Machine Vision Conference 2022, BMVC 2022, London, UK, November 21-24, 2022, BMVA Press, 2022.

[31] Shuran Song, Samuel P. Lichtenberg, Jianxiong Xiao, Sun rgb-d: A rgb-d scene understanding benchmark suite, in: Proceedings of the IEEE Conference on Computer Vision and Pattern Recognition, 2015, pp. 567–576.

[32] Mike Roberts, Jason Ramapuram, Anurag Ranjan, Atulit Kumar, Miguel Angel Bautista, Nathan Paczan, Russ Webb, Joshua M. Susskind, Hypersim: A photorealistic synthetic dataset for holistic indoor scene understanding, in: Proceedings of the IEEE/CVF International Conference on Computer Vision, 2021, pp. 10912–10922.

[33] Yunkang Cao, Xiaohao Xu, Weiming Shen, Complementary pseudo multimodal feature for point cloud anomaly detection, 2023.

[34] Jiaqi Liu, Guoyang Xie, Xinpeng Li, Jinbao Wang, Yong Liu, Chengjie Wang, Feng Zheng, et al., Real3AD: A dataset of point cloud anomaly detection, in: Thirty-Seventh Conference on Neural Information Processing Systems Datasets and Benchmarks Track, 2023.

[35] Hang Su, Subhransu Maji, Evangelos Kalogerakis, Erik Learned-Miller, Multi-view convolutional neural networks for 3d shape recognition, in: Proceedings of the IEEE International Conference on Computer Vision, 2015, pp. 945–953.

[36] Boxsang Koo, Raekyu Jung, Youngsu Yu, Inhan Kim, A geometric deep learning approach for checking element-to-entity mappings in infrastructure building information models, J. Comput. Des. Eng. 8 (1) (2021) 239–250.

[37] Charles R. Qi, Hao Su, Kaichun Mo, Leonidas J. Guibas, Pointnet: Deep learning on point sets for 3d classification and segmentation, in: Proceedings of the IEEE Conference on Computer Vision and Pattern Recognition, 2017, pp. 652–660.

[38] Zifan Shao, Kuangrong Hao, Bing Wei, Xue-Song Tang, Solder joint defect detection based on depth image CNN for 3D shape classification, in: 2021 CAA Symposium on Fault Detection, Supervision, and Safety for Technical Processes, SAFEPROCESS, IEEE, 2021, pp. 1–6.

[39] Tan Yu, Jingjing Meng, Junsong Yuan, Multi-view harmonized bilinear network for 3d object recognition, in: Proceedings of the IEEE Conference on Computer Vision and Pattern Recognition, 2018, pp. 186–194.

[40] Chao Ma, Yulan Guo, Jungang Yang, Wei An, Learning multi-view representation with LSTM for 3-D shape recognition and retrieval, IEEE Trans. Multimed. 21 (5) (2018) 1169–1182.

[41] Zhizhong Han, Mingyang Shang, Zhenbao Liu, Chi-Man Vong, Yu-Shen Liu, Matthias Zwicker, Junwei Han, C.L. Philip Chen, SeqViews2SeqLabels: Learning 3D global features via aggregating sequential views by RNN with attention, IEEE Trans. Image Process. 28 (2) (2018) 658–672.

[42] Zhizhong Han, Honglei Lu, Zhenbao Liu, Chi-Man Vong, Yu-Shen Liu, Matthias Zwicker, Junwei Han, C.L. Philip Chen, 3D2SeqViews: Aggregating sequential views for 3D global feature learning by CNN with hierarchical attention aggregation, IEEE Trans. Image Process. 28 (8) (2019) 3986–3999.

[43] Yinan Wang, Wenbo Sun, Jionghua Jin, Zhenyu Kong, Xiaowei Yue, MVGCN: Multi-view graph convolutional neural network for surface defect identification using three-dimensional point cloud, J. Manuf. Sci. Eng. 145 (3) (2023) 031004.

[44] Xin Wei, Ruixuan Yu, Jian Sun, View-gcn: View-based graph convolutional network for 3d shape analysis, in: Proceedings of the IEEE/CVF Conference on Computer Vision and Pattern Recognition, 2020, pp. 1850–1859.

[45] Qi Liang, Qiang Li, Lihu Zhang, Haixiao Mi, Weizhi Nie, Xuanya Li, MHFP: Multi-view based hierarchical fusion pooling method for 3D shape recognition, Pattern Recognit. Lett. 150 (2021) 214–220.

[46] Wenju Wang, Xiaolin Wang, Gang Chen, Haoran Zhou, Multi-view SoftPool attention convolutional networks for 3D model classification, Front. Neurorobot. 16 (2022) 1029968.

[47] Daniel Maturana, Sebastian Scherer, Voxnet: A 3d convolutional neural network for real-time object recognition, in: 2015 IEEE/RSJ International Conference on Intelligent Robots and Systems, IROS, IEEE, 2015, pp. 922–928.

[48] Yangyan Li, Soeren Pirk, Hao Su, Charles R. Qi, Leonidas J. Guibas, Fpnn: Field probing neural networks for 3d data, Adv. Neural Inf. Process. Syst. 29 (2016).

[49] Gernot Riegler, Ali Osman Ulusoy, Andreas Geiger, Octnet: Learning deep 3d representations at high resolutions, in: Proceedings of the IEEE Conference on Computer Vision and Pattern Recognition, 2017, pp. 3577–3586.

[50] Roman Klokov, Victor Lempitsky, Escape from cells: Deep kd-networks for the recognition of 3d point cloud models, in: Proceedings of the IEEE International Conference on Computer Vision, 2017, pp. 863–872.

[51] Yizhak Ben-Shabat, Michael Lindenbaum, Anath Fischer, 3Dmfv: Three-dimensional point cloud classification in real-time using convolutional neural networks, IEEE Robot. Autom. Lett. 3 (4) (2018) 3145–3152.

[52] Truc Le, Ye Duan, Pointgrid: A deep network for 3d shape understanding, in: Proceedings of the IEEE Conference on Computer Vision and Pattern Recognition, 2018, pp. 9204–9214.

[53] A.A.M. Muzahid, Wanggen Wan, Ferdous Sohel, Naimat Ullah Khan, Ofelia Delfina Cervantes Villagómez, Hidayat Ullah, 3D object classification using a volumetric deep neural network: An efficient octree guided auxiliary learning approach, IEEE Access 8 (2020) 23802–23816.

[54] Charles Ruizhongtai Qi, Li Yi, Hao Su, Leonidas J. Guibas, Pointnet++: Deep hierarchical feature learning on point sets in a metric space, Adv. Neural Inf. Process. Syst. 30 (2017).

[55] Majid Nasrollahi, Neshat Bolourian, Amin Hammad, Concrete surface defect detection using deep neural network based on lidar scanning, in: Proceedings of the CSCE Annual Conference, Laval, Greater Montreal, QC, Canada, 2019, pp. 12–15.

[56] Jin-Man Park, Yong-Ho Yoo, Ue-Hwan Kim, Dukyoung Lee, Jong-Hwan Kim, D 3 pointnet: Dual-level defect detection pointnet for solder paste printer in surface mount technology, IEEE Access 8 (2020) 140310–140322.

[57] Jiaxin Li, Ben M. Chen, Gim Hee Lee, So-net: Self-organizing network for point cloud analysis, in: Proceedings of the IEEE Conference on Computer Vision and Pattern Recognition, 2018, pp. 9397–9406.

[58] Guocheng Qian, Yuchen Li, Houwen Peng, Jinjie Mai, Hasan Abed Al Kader Hammoud, Mohamed Elhoseiny, Bernard Ghanem, PointNeXt: Revisiting PointNet++ with improved training and scaling strategies, 2022.

[59] Mor Joseph-Rivlin, Alon Zvirin, Ron Kimmel, Momen (e) t: Flavor the moments in learning to classify shapes, in: Proceedings of the IEEE/CVF International Conference on Computer Vision Workshops, 2019.

[60] Jiancheng Yang, Qiang Zhang, Bingbing Ni, Linguo Li, Jinxian Liu, Mengdie Zhou, Qi Tian, Modeling point clouds with self-attention and gumbel subset sampling, in: Proceedings of the IEEE/CVF Conference on Computer Vision and Pattern Recognition, 2019, pp. 3323–3332.

[61] Hengshuang Zhao, Li Jiang, Chi-Wing Fu, Jiaya Jia, Pointweb: Enhancing local neighborhood features for point cloud processing, in: Proceedings of the IEEE/CVF Conference on Computer Vision and Pattern Recognition, 2019, pp. 5565–5573.

[62] Juan Du, Hao Yan, Tzyy-Shuh Chang, Jianjun Shi, A tensor voting-based surface anomaly classification approach by using 3D point cloud data, J. Manuf. Sci. Eng. 144 (5) (2022) 051005.

[63] Xiao Sun, Zhouhui Lian, Jianguo Xiao, Srinet: Learning strictly rotation-invariant representations for point cloud classification and segmentation, in: Proceedings of the 27th ACM International Conference on Multimedia, 2019, pp. 980–988.

[64] Xu Yan, Chaoda Zheng, Zhen Li, Sheng Wang, Shuguang Cui, Pointasnl: Robust point clouds processing using nonlocal neural networks with adaptive sampling, in: Proceedings of the IEEE/CVF Conference on Computer Vision and Pattern Recognition, 2020, pp. 5589–5598.

[65] Amir Hertz, Rana Hanocka, Raja Giryes, Daniel Cohen-Or, Pointgmm: A neural gmm network for point clouds, in: Proceedings of the IEEE/CVF Conference on Computer Vision and Pattern Recognition, 2020, pp. 12054–12063.

[66] Nianniang Wang, Duo Ma, Xueming Du, Bin Li, Danyang Di, Gaozhao Pang, Yihang Duan, An automatic defect classification and segmentation method on three-dimensional point clouds for sewer pipes, Tunn. Undergr. Space Technol. 143 (2024) 105480.

[67] Meng-Hao Guo, Jun-Xiong Cai, Zheng-Ning Liu, Tai-Jiang Mu, Ralph R. Martin, Shi-Min Hu, Pct: Point cloud transformer, Comput. Vis. Media 7 (2021) 187–199.

[68] Joakim Bruslund Haurum, Moaaz M.J. Allahham, Mathias S. Lynge, Kasper Schøn Henriksen, Ivan A. Nikolov, Thomas B. Moeslund, Sewer defect classification using synthetic point clouds, in: VISIGRAPP (5: VISAPP), 2021, pp. 891–900.

[69] Yunxiang Zhou, Ankang Ji, Limao Zhang, Sewer defect detection from 3D point clouds using a transformer-based deep learning model, Autom. Constr. 136 (2022) 104163.

[70] Varun Kasireddy, Burcu Akinci, Encoding 3D point contexts for self-supervised spall classification using 3D bridge point clouds, J. Comput. Civ. Eng. 37 (2) (2023) 04022061.

[71] Christian Benz, Volker Rodehorst, Multi-view 3D instance segmentation of structural anomalies for enhanced structural inspection of concrete bridges, 2024, arXiv preprint arXiv:2401.03298.

[72] Hengshuang Zhao, Li Jiang, Jiaya Jia, Philip Torr, Vladlen Koltun, Point transformer, 2021.

[73] Luyao Liu, Enqing Chen, Yingqiang Ding, TR-Net: a transformer-based neural network for point cloud processing, Machines 10 (7) (2022) 517.

[74] Xumin Yu, Lulu Tang, Yongming Rao, Tiejun Huang, Jie Zhou, Jiwen Lu, Point-bert: Pre-training 3d point cloud transformers with masked point modeling, in: Proceedings of the IEEE/CVF Conference on Computer Vision and Pattern Recognition, 2022, pp. 19313–19322.

[75] Yatian Pang, Wenxiao Wang, Francis E.H. Tay, Wei Liu, Yonghong Tian, Li Yuan, Masked autoencoders for point cloud self-supervised learning, in: European Conference on Computer Vision, Springer, 2022, pp. 604–621.

[76] Yahui Liu, Bin Tian, Yisheng Lv, Lingxi Li, Fei-Yue Wang, Point cloud classification using content-based transformer via clustering in feature space, IEEE/CAA J. Autom. Sin. (2023).

[77] Lequn Chen, Xiling Yao, Peng Xu, Seung Ki Moon, Guijun Bi, Rapid surface defect identification for additive manufacturing with in-situ point cloud processing and machine learning, Virtual Phys. Prototyp. 16 (1) (2021) 50–67.

[78] Rui Li, Mingzhou Jin, Vincent C. Paquit, Geometrical defect detection for additive manufacturing with machine learning models, Mater. Des. 206 (2021) 109726.

[79] Alex Krizhevsky, Ilya Sutskever, Geoffrey E. Hinton, Imagenet classification with deep convolutional neural networks, Adv. Neural Inf. Process. Syst. 25 (2012).

[80] Yann LeCun, Bernhard Boser, John Denker, Donnie Henderson, Richard Howard, Wayne Hubbard, Lawrence Jackel, Handwritten digit recognition with a back-propagation network, Adv. Neural Inf. Process. Syst. 2 (1989).

[81] Pedro Hermosilla, Tobias Ritschel, Pere-Pau Vázquez, Àlvar Vinacua, Timo Ropinski, Monte carlo convolution for learning on non-uniformly sampled point clouds, ACM Trans. Graph. 37 (6) (2018) 1–12.

[82] Yifan Xu, Tianqi Fan, Mingye Xu, Long Zeng, Yu Qiao, Spidercnn: Deep learning on point sets with parameterized convolutional filters, in: Proceedings of the European Conference on Computer Vision, ECCV, 2018, pp. 87–102.

[83] Alexandre Boulch, Gilles Puy, Renaud Marlet, FKAConv: Feature-kernel alignment for point cloud convolution, in: Proceedings of the Asian Conference on Computer Vision, 2020.

[84] Wenxuan Wu, Zhongang Qi, Li Fuxin, Pointconv: Deep convolutional networks on 3d point clouds, in: Proceedings of the IEEE/CVF Conference on Computer Vision and Pattern Recognition, 2019, pp. 9621–9630.

[85] Hugues Thomas, Charles R. Qi, Jean-Emmanuel Deschaud, Beatriz Marcotegui, François Goulette, Leonidas J. Guibas, Kpconv: Flexible and deformable convolution for point clouds, in: Proceedings of the IEEE/CVF International Conference on Computer Vision, 2019, pp. 6411–6420.

[86] Alexandre Boulch, ConvPoint: Continuous convolutions for point cloud processing, Comput. Graph. 88 (2020) 24–34.

[87] Yangyan Li, Rui Bu, Mingchao Sun, Wei Wu, Xinhan Di, Baoquan Chen, Pointcnn: Convolution on x-transformed points, Adv. Neural Inf. Process. Syst. 31 (2018).

[88] Binh-Son Hua, Minh-Khoi Tran, Sai-Kit Yeung, Pointwise convolutional neural networks, in: Proceedings of the IEEE Conference on Computer Vision and Pattern Recognition, 2018, pp. 984–993.

[89] Adrien Poulenard, Marie-Julie Rakotosaona, Yann Ponty, Maks Ovsjanikov, Effective rotation-invariant point cnn with spherical harmonics kernels, in: 2019 International Conference on 3D Vision, 3DV, IEEE, 2019, pp. 47–56.

[90] Changshuo Wang, Xin Ning, Linjun Sun, Liping Zhang, Weijun Li, Xiao Bai, Learning discriminative features by covering local geometric space for point cloud analysis, IEEE Trans. Geosci. Remote Sens. 60 (2022) 1–15.

[91] Martin Simonovsky, Nikos Komodakis, Dynamic edge-conditioned filters in convolutional neural networks on graphs, in: Proceedings of the IEEE Conference on Computer Vision and Pattern Recognition, 2017, pp. 3693–3702.

[92] Yingxue Zhang, Michael Rabbat, A graph-cnn for 3d point cloud classification, in: 2018 IEEE International Conference on Acoustics, Speech and Signal Processing, ICASSP, IEEE, 2018, pp. 6279–6283.

[93] Jinxian Liu, Bingbing Ni, Caiyuan Li, Jiancheng Yang, Qi Tian, Dynamic points agglomeration for hierarchical point sets learning, in: Proceedings of the IEEE/CVF International Conference on Computer Vision, 2019, pp. 7546–7555.

[94] Kuangen Zhang, Ming Hao, Jing Wang, Xinxing Chen, Yuquan Leng, Clarence W. de Silva, Chenglong Fu, Linked dynamic graph cnn: Learning through point cloud by linking hierarchical features, in: 2021 27th International Conference on Mechatronics and Machine Vision in Practice, M2VIP, IEEE, 2021, pp. 7–12.

[95] Seyed Saber Mohammadi, Yiming Wang, Alessio Del Bue, Pointview-gcn: 3d shape classification with multi-view point clouds, in: 2021 IEEE International Conference on Image Processing, ICIP, IEEE, 2021, pp. 3103–3107.

[96] Long Hoang, Suk-Hwan Lee, Eung-Joo Lee, Ki-Ryong Kwon, GSV-NET: A multi-modal deep learning network for 3D point cloud classification, Appl. Sci. 12 (1) (2022) 483.

[97] Lifang Chen, Qian Zhang, DDGCN: graph convolution network based on direction and distance for point cloud learning, Vis. Comput. 39 (3) (2023) 863–873.

[98] Jiaying Zhang, Xiaoli Zhao, Zheng Chen, Zhejun Lu, A review of deep learning-based semantic segmentation for point cloud, IEEE Access 7 (2019) 179118–179133.

[99] Alok Jhaldiyal, Navendu Chaudhary, Semantic segmentation of 3D LiDAR data using deep learning: a review of projection-based methods, Appl. Intell. 53 (6) (2023) 6844–6855.

[100] Tian Xia, Jian Yang, Long Chen, Automated semantic segmentation of bridge point cloud based on local descriptor and machine learning, Autom. Constr. 133 (2022) 103992.

[101] Felix Järemo Lawin, Martin Danelljan, Patrik Tosteberg, Goutam Bhat, Fahad Shahbaz Khan, Michael Felsberg, Deep projective 3D semantic segmentation, in: Computer Analysis of Images and Patterns: 17th International Conference, CAIP 2017, Ystad, Sweden, August 22-24, 2017, Proceedings, Part I 17, Springer, 2017, pp. 95–107.

[102] Alexandre Boulch, Bertrand Le Saux, Nicolas Audebert, et al., Unstructured point cloud semantic labeling using deep segmentation networks, in: 3dor@ Eurographics, Vol. 3, 2017, pp. 17–24.

[103] Vijay Badrinarayanan, Alex Kendall, Roberto Cipolla, Segnet: A deep convolutional encoder-decoder architecture for image segmentation, IEEE Trans. Pattern Anal. Mach. Intell. 39 (12) (2017) 2481–2495.

[104] Olaf Ronneberger, Philipp Fischer, Thomas Brox, U-net: Convolutional networks for biomedical image segmentation, in: Medical Image Computing and Computer-Assisted Intervention–MICCAI 2015: 18th International Conference, Munich, Germany, October 5-9, 2015, Proceedings, Part III 18, Springer, 2015, pp. 234–241.

[105] Nicolas Audebert, Bertrand Le Saux, Sébastien Lefèvre, Semantic segmentation of earth observation data using multimodal and multi-scale deep networks, in: Computer Vision–ACCV 2016: 13th Asian Conference on Computer Vision, Taipei, Taiwan, November 20-24, 2016, Revised Selected Papers, Part I 13, Springer, 2017, pp. 180–196.

[106] Alexandre Boulch, Joris Guerry, Bertrand Le Saux, Nicolas Audebert, SnapNet: 3D point cloud semantic labeling with 2D deep segmentation networks, Comput. Graph. 71 (2018) 189–198.

[107] Maxim Tatarchenko, Jaesik Park, Vladlen Koltun, Qian-Yi Zhou, Tangent convolutions for dense prediction in 3d, in: Proceedings of the IEEE Conference on Computer Vision and Pattern Recognition, 2018, pp. 3887–3896.

[108] Haibo Qiu, Baosheng Yu, Dacheng Tao, Gfnet: Geometric flow network for 3d point cloud semantic segmentation, 2022, arXiv preprint arXiv:2207.02605.

[109] Qihang Wang, Xiaoming Wang, Qing He, Jun Huang, Hong Huang, Ping Wang, Tianle Yu, Min Zhang, 3D tensor-based point cloud and image fusion for robust detection and measurement of rail surface defects, Autom. Constr. 161 (2024) 105342.

[110] Bichen Wu, Alvin Wan, Xiangyu Yue, Kurt Keutzer, Squeezeseg: Convolutional neural nets with recurrent crf for real-time road-object segmentation from 3d lidar point cloud, in: 2018 IEEE International Conference on Robotics and Automation, ICRA, IEEE, 2018, pp. 1887–1893.

[111] Bichen Wu, Xuanyu Zhou, Sicheng Zhao, Xiangyu Yue, Kurt Keutzer, Squeezesegv2: Improved model structure and unsupervised domain adaptation for road-object segmentation from a lidar point cloud, in: 2019 International Conference on Robotics and Automation, ICRA, IEEE, 2019, pp. 4376–4382.

[112] Andres Milioto, Ignacio Vizzo, Jens Behley, Cyrill Stachniss, Rangenet++: Fast and accurate lidar semantic segmentation, in: 2019 IEEE/RSJ International Conference on Intelligent Robots and Systems, IROS, IEEE, 2019, pp. 4213–4220.

[113] Joseph Redmon, Ali Farhadi, Yolov3: An incremental improvement, 2018, arXiv preprint arXiv:1804.02767.

[114] Jinhui Zhang, Haiyan Jiang, Huizhi Shao, Qingjun Song, Xiaoshuang Wang, Dashuai Zong, Semantic segmentation of in-vehicle point cloud with improved RangeNet++ loss function, IEEE Access 11 (2023) 8569–8580.

[115] Yin Zhou, Oncel Tuzel, Voxelnet: End-to-end learning for point cloud based 3d object detection, in: Proceedings of the IEEE Conference on Computer Vision and Pattern Recognition, 2018, pp. 4490–4499.

[116] Jing Huang, Suya You, Point cloud labeling using 3d convolutional neural network, in: 2016 23rd International Conference on Pattern Recognition, ICPR, IEEE, 2016, pp. 2670–2675.

[117] Mehrdad S. Dizaji, Devin K. Harris, 3D InspectionNet: a deep 3D convolutional neural networks based approach for 3D defect detection on concrete columns, in: Nondestructive Characterization and Monitoring of Advanced Materials, Aerospace, Civil Infrastructure, and Transportation XIII, Vol. 10971, SPIE, 2019, pp. 67–77.

[118] Lyne Tchapmi, Christopher Choy, Iro Armeni, JunYoung Gwak, Silvio Savarese, Segcloud: Semantic segmentation of 3d point clouds, in: 2017 International Conference on 3D Vision, 3DV, IEEE, 2017, pp. 537–547.

[119] Hsien-Yu Meng, Lin Gao, Yu-Kun Lai, Dinesh Manocha, Vv-net: Voxel vae net with group convolutions for point cloud segmentation, in: Proceedings of the IEEE/CVF International Conference on Computer Vision, 2019, pp. 8500–8508.

[120] Eren Erdal Aksoy, Saimir Baci, Selcuk Cavdar, Salsanet: Fast road and vehicle segmentation in lidar point cloud for autonomous driving, in: 2020 IEEE Intelligent Vehicles Symposium, IV, IEEE, 2020, pp. 926–932.

[121] Tiago Cortinhal, George Tzelepis, Eren Erdal Aksoy, Salsanext: Fast, uncertainty-aware semantic segmentation of lidar point clouds, in: Advances in Visual Computing: 15th International Symposium, ISVC 2020, San Diego, CA, USA, October 5–7, 2020, Proceedings, Part II 15, Springer, 2020, pp. 207–222.

[122] Ran Cheng, Ryan Razani, Ehsan Taghavi, Enxu Li, Bingbing Liu, 2-s3net: Attentive feature fusion with adaptive feature selection for sparse semantic segmentation network, in: Proceedings of the IEEE/CVF Conference on Computer Vision and Pattern Recognition, 2021, pp. 12547–12556.

[123] Ran Cheng, Ryan Razani, Yuan Ren, Liu Bingbing, S3Net: 3D LiDAR sparse semantic segmentation network, in: 2021 IEEE International Conference on Robotics and Automation, ICRA, IEEE, 2021, pp. 14040–14046.

[124] Ran Cheng, Christopher Agia, Yuan Ren, Xinhai Li, Liu Bingbing, S3cnet: A sparse semantic scene completion network for lidar point clouds, in: Conference on Robot Learning, PMLR, 2021, pp. 2148–2161.

[125] Xiaoyan Li, Gang Zhang, Hongyu Pan, Zhenhua Wang, CPGNet: Cascade point-grid fusion network for real-time LiDAR semantic segmentation, in: 2022 International Conference on Robotics and Automation, ICRA, IEEE, 2022, pp. 11117–11123.

[126] Jaehyun Park, Chansoo Kim, Soyeong Kim, Kichun Jo, PCSCNet: Fast 3D semantic segmentation of LiDAR point cloud for autonomous car using point convolution and sparse convolution network, Expert Syst. Appl. 212 (2023) 118815.

[127] Miguel Martin-Abadal, Manuel Piñar-Molina, Antoni Martorell-Torres, Gabriel Oliver-Codina, Yolanda Gonzalez-Cid, Underwater pipe and valve 3D recognition using deep learning segmentation, J. Mar. Sci. Eng. 9 (1) (2020) 5.

[128] Neshat Bolourian, Majid Nasrollahi, Fardin Bahreini, Amin Hammad, Point cloud–based concrete surface defect semantic segmentation, J. Comput. Civ. Eng. 37 (2) (2023) 04022056.

[129] Jiangpeng Shu, Wenhao Li, Congguang Zhang, Yifan Gao, Yiqiang Xiang, Ling Ma, Point cloud-based dimensional quality assessment of precast concrete components using deep learning, J. Build. Eng. 70 (2023) 106391.

[130] Chao Yin, Boyu Wang, Vincent J.L. Gan, Mingzhu Wang, Jack C.P. Cheng, Automated semantic segmentation of industrial point clouds using ResPointNet++, Autom. Constr. 130 (2021) 103874.

[131] Mingyang Jiang, Yiran Wu, Tianqi Zhao, Zelin Zhao, Cewu Lu, Pointsift: A siftlike network module for 3d point cloud semantic segmentation, 2018, arXiv preprint arXiv:1807.00652.

[132] Francis Engelmann, Theodora Kontogianni, Jonas Schult, Bastian Leibe, Know what your neighbors do: 3D semantic segmentation of point clouds, in: Proceedings of the European Conference on Computer Vision (ECCV) Workshops, 2018.

[133] Qingyong Hu, Bo Yang, Linhai Xie, Stefano Rosa, Yulan Guo, Zhihua Wang, Niki Trigoni, Andrew Markham, Randla-net: Efficient semantic segmentation of large-scale point clouds, in: Proceedings of the IEEE/CVF Conference on Computer Vision and Pattern Recognition, 2020, pp. 11108–11117.

[134] Chuanyu Luo, Xiaohan Li, Nuo Cheng, Han Li, Shengguang Lei, Pu Li, MVPnet: Multiple view pointwise semantic segmentation of large-scale point clouds, 2022, arXiv preprint arXiv:2201.12769.

[135] Varun Kasireddy, Burcu Akinci, Assessing the impact of 3D point neighborhood size selection on unsupervised spall classification with 3D bridge point clouds, Adv. Eng. Inform. 52 (2022) 101624.

[136] Lin-Zhuo Chen, Xuan-Yi Li, Deng-Ping Fan, Kai Wang, Shao-Ping Lu, Ming-Ming Cheng, LSANet: Feature learning on point sets by local spatial aware layer, 2019, arXiv preprint arXiv:1905.05442.

[137] Chenxi Zhao, Weihao Zhou, Li Lu, Qijun Zhao, Pooling scores of neighboring points for improved 3D point cloud segmentation, in: 2019 IEEE International Conference on Image Processing, ICIP, IEEE, 2019, pp. 1475–1479.

[138] Zhiyu Hu, Dongbo Zhang, Shuai Li, Hong Qin, Attention-based relation and context modeling for point cloud semantic segmentation, Comput. Graph. 90 (2020) 126–134.

[139] Shuang Deng, Qiulei Dong, GA-NET: Global attention network for point cloud semantic segmentation, IEEE Signal Process. Lett. 28 (2021) 1300–1304.

[140] Ankang Ji, Yunxiang Zhou, Limao Zhang, Robert L.K. Tiong, Xiaolong Xue, Semi-supervised learning-based point cloud network for segmentation of 3D tunnel scenes, Autom. Constr. 146 (2023) 104668.

[141] Yunxiang Zhou, Ankang Ji, Limao Zhang, Xiaolong Xue, Attention-enhanced sampling point cloud network (ASPCNet) for efficient 3D tunnel semantic segmentation, Autom. Constr. 146 (2023) 104667.

[142] Jaesung Choe, Chunghyun Park, Francois Rameau, Jaesik Park, In So Kweon, Pointmixer: Mlp-mixer for point cloud understanding, in: European Conference on Computer Vision, Springer, 2022, pp. 620–640.

[143] Na Zhao, Tat-Seng Chua, Gim Hee Lee, Ps²2-net: A locally and globally aware network for point-based semantic segmentation, in: 2020 25th International Conference on Pattern Recognition, ICPR, IEEE, 2021, pp. 723–730.

[144] Relja Arandjelovic, Petr Gronat, Akihiko Torii, Tomas Pajdla, Josef Sivic, NetVLAD: CNN architecture for weakly supervised place recognition, in: Proceedings of the IEEE Conference on Computer Vision and Pattern Recognition, 2016, pp. 5297–5307.

[145] Siqi Fan, Qiulei Dong, Fenghua Zhu, Yisheng Lv, Peijun Ye, Fei-Yue Wang, SCF-Net: Learning spatial contextual features for large-scale point cloud segmentation, in: Proceedings of the IEEE/CVF Conference on Computer Vision and Pattern Recognition, 2021, pp. 14504–14513.

[146] Yuanwei Bi, Lujian Zhang, Yaowen Liu, Yansen Huang, Hao Liu, A local-global feature fusing method for point clouds semantic segmentation, IEEE Access (2023).

[147] Yiqiang Zhao, Xingyi Ma, Bin Hu, Qi Zhang, Mao Ye, Guoqing Zhou, A large-scale point cloud semantic segmentation network via local dual features and global correlations, Comput. Graph. 111 (2023) 133–144.

[148] Xiaohan Tu, Chuanhao Zhang, Siping Liu, Cheng Xu, Renfa Li, Point cloud segmentation of overhead contact systems with deep learning in high-speed rails, J. Netw. Comput. Appl. 216 (2023) 103671.

[149] Anh-Thuan Tran, Hoanh-Su Le, Suk-Hwan Lee, Ki-Ryong Kwon, PointCT: Point central transformer network for weakly-supervised point cloud semantic segmentation, in: Proceedings of the IEEE/CVF Winter Conference on Applications of Computer Vision, 2024, pp. 3556–3565.

[150] Junfeng Jing, Huaqing Wang, Defect segmentation with local embedding in industrial 3D point clouds based on transformer, Meas. Sci. Technol. 35 (3) (2023) 035406.

[151] Yeritza Perez-Perez, Mani Golparvar-Fard, Khaled El-Rayes, Scan2BIM-NET: Deep learning method for segmentation of point clouds for scan-to-BIM, J. Constr. Eng. Manag. 147 (9) (2021) 04021107.

[152] Zhenxing Xu, Aizeng Wang, Fei Hou, Gang Zhao, Defect detection of gear parts in virtual manufacturing, Vis. Comput. Ind. Biomed. Art 6 (1) (2023) 1–12.

[153] Shenlong Wang, Simon Suo, Wei-Chiu Ma, Andrei Pokrovsky, Raquel Urtasun, Deep parametric continuous convolutional neural networks, in: Proceedings of the IEEE Conference on Computer Vision and Pattern Recognition, 2018, pp. 2589–2597.

[154] Yong Li, Xu Li, Zhenxin Zhang, Feng Shuang, Qi Lin, Jincheng Jiang, DenseKP-NET: Dense kernel point convolutional neural networks for point cloud semantic segmentation, IEEE Trans. Geosci. Remote Sens. 60 (2022) 1–13.

[155] Guofeng Tong, Yuyuan Shao, Hao Peng, Learning local contextual features for 3D point clouds semantic segmentation by attentive kernel convolution, Vis. Comput. (2023) 1–17.

[156] Francis Engelmann, Theodora Kontogianni, Bastian Leibe, Dilated point convolutions: On the receptive field size of point convolutions on 3d point clouds, in: 2020 IEEE International Conference on Robotics and Automation, ICRA, IEEE, 2020, pp. 9463–9469.

[157] Wenxuan Wu, Li Fuxin, Qi Shan, Pointconvformer: Revenge of the point-based convolution, in: Proceedings of the IEEE/CVF Conference on Computer Vision and Pattern Recognition, 2023, pp. 21802–21813.

[158] Xinyue Zhao, Quanzhi Li, Menghan Xiao, Zaixing He, Defect detection of 3D printing surface based on geometric local domain features, Int. J. Adv. Manuf. Technol. 125 (1) (2023) 183–194.

[159] Mengru Liu, Xingwang Bai, Shengxuan Xi, Honghui Dong, Runsheng Li, Haiou Zhang, Xiangman Zhou, Detection and quantitative evaluation of surface defects in wire and arc additive manufacturing based on 3D point cloud, Virtual Phys. Prototyp. 19 (1) (2024) e2294336.

[160] F. Bahreini, A. Hammad, Point cloud semantic segmentation of concrete surface defects using dynamic graph CNN, in: ISARC. Proceedings of the International Symposium on Automation and Robotics in Construction, Vol. 38, IAARC Publications, 2021, pp. 379–386.

[161] Hyeonsoo Kim, Changwan Kim, Deep-learning-based classification of point clouds for bridge inspection, Remote Sens. 12 (22) (2020) 3757.

[162] Roberto Pierdicca, Marina Paolanti, Francesca Matrone, Massimo Martini, Christian Morbidoni, Eva Savina Malinverni, Emanuele Frontoni, Andrea Maria Lingua, Point cloud semantic segmentation using a deep learning framework for cultural heritage, Remote Sens. 12 (6) (2020) 1005.

[163] Jie Nie, Lei Huang, Chengyu Zheng, Xiaowei Lv, Rui Wang, Cross-scale graph interaction network for semantic segmentation of remote sensing images, ACM Trans. Multimed. Comput. Commun. Appl. 19 (6) (2023) 1–18.

[164] Chengzhi Wu, Xuelei Bi, Julius Pfrommer, Alexander Cebulla, Simon Mangold, Jürgen Beyerer, Sim2real transfer learning for point cloud segmentation: An industrial application case on autonomous disassembly, in: Proceedings of the IEEE/CVF Winter Conference on Applications of Computer Vision, 2023, pp. 4531–4540.

[165] Mingtao Feng, Liang Zhang, Xuefei Lin, Syed Zulqarnain Gilani, Ajmal Mian, Point attention network for semantic segmentation of 3D point clouds, Pattern Recognit. 107 (2020) 107446.

[166] Zijin Du, Hailiang Ye, Feilong Cao, A novel local-global graph convolutional method for point cloud semantic segmentation, IEEE Trans. Neural Netw. Learn. Syst. (2022).

[167] Helin Li, Bin Lin, Chen Zhang, Liang Xu, Tianyi Sui, Yang Wang, Xinquan Hao, Deyu Lou, Hongyu Li, Mask-Point: automatic 3D surface defects detection network for fiber-reinforced resin matrix composites, Polymers 14 (16) (2022) 3390.

[168] In-Ho Kim, Haemin Jeon, Seung-Chan Baek, Won-Hwa Hong, Hyung-Jo Jung, Application of crack identification techniques for an aging concrete bridge inspection using an unmanned aerial vehicle, Sensors 18 (6) (2018) 1881.

[169] Ji Hou, Angela Dai, Matthias Nießner, 3D-sis: 3d semantic instance segmentation of rgb-d scans, in: Proceedings of the IEEE/CVF Conference on Computer Vision and Pattern Recognition, 2019, pp. 4421–4430.

[170] Jinhua Lin, Lin Ma, Yu Yao, Segmentation of casting defect regions for the extraction of microstructural properties, Eng. Appl. Artif. Intell. 85 (2019) 150–163.

[171] Li Yi, Wang Zhao, He Wang, Minhyuk Sung, Leonidas J. Guibas, Gspn: Generative shape proposal network for 3d instance segmentation in point cloud, in: Proceedings of the IEEE/CVF Conference on Computer Vision and Pattern Recognition, 2019, pp. 3947–3956.

[172] Bo Yang, Jianan Wang, Ronald Clark, Qingyong Hu, Sen Wang, Andrew Markham, Niki Trigoni, Learning object bounding boxes for 3d instance segmentation on point clouds, Adv. Neural Inf. Process. Syst. 32 (2019).

[173] Shih-Hung Liu, Shang-Yi Yu, Shao-Chi Wu, Hwann-Tzong Chen, Tyng-Luh Liu, Learning gaussian instance segmentation in point clouds, 2020, arXiv preprint arXiv:2007.09860.

[174] Lei Han, Tian Zheng, Lan Xu, Lu Fang, Occuseg: Occupancy-aware 3d instance segmentation, in: Proceedings of the IEEE/CVF Conference on Computer Vision and Pattern Recognition, 2020, pp. 2940–2949.

[175] Ze Liu, Zheng Zhang, Yue Cao, Han Hu, Xin Tong, Group-free 3d object detection via transformers, in: Proceedings of the IEEE/CVF International Conference on Computer Vision, 2021, pp. 2949–2958.

[176] Ruohao Guo, Dantong Niu, Liao Qu, Zhenbo Li, Sotr: Segmenting objects with transformers, in: Proceedings of the IEEE/CVF International Conference on Computer Vision, 2021, pp. 7157–7166.

[177] Yutong Xie, Jianpeng Zhang, Chunhua Shen, Yong Xia, Cotr: Efficiently bridging cnn and transformer for 3d medical image segmentation, in: Medical Image Computing and Computer Assisted Intervention–MICCAI 2021: 24th International Conference, Strasbourg, France, September 27–October 1, 2021, Proceedings, Part III 24, Springer, 2021, pp. 171–180.

[178] Justin Lazarow, Weijian Xu, Zhuowen Tu, Instance segmentation with mask-supervised polygonal boundary transformers, in: Proceedings of the IEEE/CVF Conference on Computer Vision and Pattern Recognition, 2022, pp. 4382–4391.

[179] Jiahao Sun, Chunmei Qing, Junpeng Tan, Xiangmin Xu, Superpoint transformer for 3d scene instance segmentation, in: Proceedings of the AAAI Conference on Artificial Intelligence, Vol. 37, 2023, pp. 2393–2401, (2).

[180] Xu Fang, Qing Li, Jiasong Zhu, Zhipeng Chen, Dejin Zhang, Kechun Wu, Kai Ding, Qingquan Li, Sewer defect instance segmentation, localization, and 3D reconstruction for sewer floating capsule robots, Autom. Constr. 142 (2022) 104494.

[181] Jonas Schult, Francis Engelmann, Alexander Hermans, Or Litany, Siyu Tang, Bastian Leibe, Mask3d: Mask transformer for 3d semantic instance segmentation, in: 2023 IEEE International Conference on Robotics and Automation, ICRA, IEEE, 2023, pp. 8216–8223.

[182] Xin Lai, Yuhui Yuan, Ruihang Chu, Yukang Chen, Han Hu, Jiaya Jia, Mask-attention-free transformer for 3d instance segmentation, in: Proceedings of the IEEE/CVF International Conference on Computer Vision, 2023, pp. 3693–3703.

[183] Weiyue Wang, Ronald Yu, Qiangui Huang, Ulrich Neumann, Sgpn: Similarity group proposal network for 3d point cloud instance segmentation, in: Proceedings of the IEEE Conference on Computer Vision and Pattern Recognition, 2018, pp. 2569–2578.

[184] Chen Liu, Yasutaka Furukawa, Masc: Multi-scale affinity with sparse convolution for 3d instance segmentation, 2019, arXiv preprint arXiv:1902.04478.

[185] Zhidong Liang, Ming Yang, Hao Li, Chunxiang Wang, 3D instance embedding learning with a structure-aware loss function for point cloud segmentation, IEEE Robot. Autom. Lett. 5 (3) (2020) 4915–4922.

[186] Xinlong Wang, Shu Liu, Xiaoyong Shen, Chunhua Shen, Jiaya Jia, Associatively segmenting instances and semantics in point clouds, in: Proceedings of the IEEE/CVF Conference on Computer Vision and Pattern Recognition, 2019, pp. 4096–4105.

[187] Lin Zhao, Wenbing Tao, JSNet: Joint instance and semantic segmentation of 3D point clouds, in: Proceedings of the AAAI Conference on Artificial Intelligence, Vol. 34, 2020, pp. 12951–12958, (7).

[188] Lin Zhao, Wenbing Tao, Jsnet++: Dynamic filters and pointwise correlation for 3d point cloud instance and semantic segmentation, IEEE Trans. Circuits Syst. Video Technol. 33 (4) (2022) 1854–1867.

[189] Francis Engelmann, Martin Bokeloh, Alireza Fathi, Bastian Leibe, Matthias Nießner, 3D-mpa: Multi-proposal aggregation for 3d semantic instance segmentation, in: Proceedings of the IEEE/CVF Conference on Computer Vision and Pattern Recognition, 2020, pp. 9031–9040.

[190] Bryan G. Pantoja-Rosero, D. Oner, Mateusz Kozinski, Radhakrishna Achanta, Pascal Fua, Fernando Pérez-Cruz, Katrin Beyer, TOPO-Loss for continuity-preserving crack detection using deep learning, Constr. Build. Mater. 344 (2022) 128264.

[191] Fabian Isensee, Paul F. Jaeger, Simon A.A. Kohl, Jens Petersen, Klaus H. Maier-Hein, nnU-Net: a self-configuring method for deep learning-based biomedical image segmentation, Nature Methods 18 (2) (2021) 203–211.

[192] Zhihao Liang, Zhihao Li, Songcen Xu, Mingkui Tan, Kui Jia, Instance segmentation in 3D scenes using semantic superpoint tree networks, in: Proceedings of the IEEE/CVF International Conference on Computer Vision, 2021, pp. 2783–2792.

[193] Thang Vu, Kookhoi Kim, Tung M. Luu, Thanh Nguyen, Chang D. Yoo, Softgroup for 3d instance segmentation on point clouds, in: Proceedings of the IEEE/CVF Conference on Computer Vision and Pattern Recognition, 2022, pp. 2708–2717.

[194] Jean Lahoud, Bernard Ghanem, Marc Pollefeys, Martin R. Oswald, 3D instance segmentation via multi-task metric learning, in: Proceedings of the IEEE/CVF International Conference on Computer Vision, 2019, pp. 9256–9266.

[195] Li Jiang, Hengshuang Zhao, Shaoshuai Shi, Shu Liu, Chi-Wing Fu, Jiaya Jia, Pointgroup: Dual-set point grouping for 3d instance segmentation, in: Proceedings of the IEEE/CVF Conference on Computer Vision and Pattern Recognition, 2020, pp. 4867–4876.

[196] Shaoyu Chen, Jiemin Fang, Qian Zhang, Wenyu Liu, Xinggang Wang, Hierarchical aggregation for 3d instance segmentation, in: Proceedings of the IEEE/CVF International Conference on Computer Vision, 2021, pp. 15467–15476.

[197] Tong He, Chunhua Shen, Anton Van Den Hengel, Dyco3d: Robust instance segmentation of 3d point clouds through dynamic convolution, in: Proceedings of the IEEE/CVF Conference on Computer Vision and Pattern Recognition, 2021, pp. 354–363.

[198] Javier Grandio, Belen Riveiro, Daniel Lamas, Pedro Arias, Multimodal deep learning for point cloud panoptic segmentation of railway environments, Autom. Constr. 150 (2023) 104854.

[199] Zhaoxiang Zhang, Ankang Ji, Limao Zhang, Yuelei Xu, Qing Zhou, Deep learning for large-scale point cloud segmentation in tunnels considering causal inference, Autom. Constr. 152 (2023) 104915.

[200] Eva Agapaki, Ioannis Brilakis, Instance segmentation of industrial point cloud data, J. Comput. Civ. Eng. 35 (6) (2021) 04021022.

[201] Zongji Wang, Feng Lu, Voxsegnet: Volumetric cnns for semantic part segmentation of 3d shapes, IEEE Trans. Vis. Comput. Graphics 26 (9) (2019) 2919–2930.

[202] Kaichun Mo, Shilin Zhu, Angel X. Chang, Li Yi, Subarna Tripathi, Leonidas J. Guibas, Hao Su, Partnet: A large-scale benchmark for fine-grained and hierarchical part-level 3d object understanding, in: Proceedings of the IEEE/CVF Conference on Computer Vision and Pattern Recognition, 2019, pp. 909–918.

[203] Evangelos Kalogerakis, Melinos Averkiou, Subhransu Maji, Siddhartha Chaudhuri, 3D shape segmentation with projective convolutional networks, in: Proceedings of the IEEE Conference on Computer Vision and Pattern Recognition, 2017, pp. 3779–3788.

[204] Li Yi, Hao Su, Xingwen Guo, Leonidas J. Guibas, Syncspeccnn: Synchronized spectral cnn for 3d shape segmentation, in: Proceedings of the IEEE Conference on Computer Vision and Pattern Recognition, 2017, pp. 2282–2290.

[205] Pengyu Wang, Yuan Gan, Panpan Shui, Fenggen Yu, Yan Zhang, Songle Chen, Zhengxing Sun, 3D shape segmentation via shape fully convolutional networks, Comput. Graph. 76 (2018) 182–192.

[206] Chungang Zhuang, Zhe Wang, Heng Zhao, Han Ding, Semantic part segmentation method based 3D object pose estimation with RGB-D images for bin-picking, Robot. Comput.-Integr. Manuf. 68 (2021) 102086.

[207] Chenyang Zhu, Kai Xu, Siddhartha Chaudhuri, Li Yi, Leonidas J. Guibas, Hao Zhang, AdaCoSeg: Adaptive shape co-segmentation with group consistency loss, in: Proceedings of the IEEE/CVF Conference on Computer Vision and Pattern Recognition, 2020, pp. 8543–8552.

[208] Zhiqin Chen, Kangxue Yin, Matthew Fisher, Siddhartha Chaudhuri, Hao Zhang, Bae-net: Branched autoencoder for shape co-segmentation, in: Proceedings of the IEEE/CVF International Conference on Computer Vision, 2019, pp. 8490–8499.

[209] Shuailin Li, Chuyu Zhang, Xuming He, Shape-aware semi-supervised 3D semantic segmentation for medical images, in: Medical Image Computing and Computer Assisted Intervention–MICCAI 2020: 23rd International Conference, Lima, Peru, October 4–8, 2020, Proceedings, Part I 23, Springer, 2020, pp. 552–561.

[210] Panos Achlioptas, Judy Fan, Robert Hawkins, Noah Goodman, Leonidas J. Guibas, ShapeGlot: Learning language for shape differentiation, in: Proceedings of the IEEE/CVF International Conference on Computer Vision, 2019, pp. 8938–8947.

[211] Juil Koo, Ian Huang, Panos Achlioptas, Leonidas J. Guibas, Minhyuk Sung, Partglot: Learning shape part segmentation from language reference games, in: Proceedings of the IEEE/CVF Conference on Computer Vision and Pattern Recognition, 2022, pp. 16505–16514.

[212] Zhenyu Shu, Sipeng Yang, Haoyu Wu, Shiqing Xin, Chaoyi Pang, Ladislav Kavan, Ligang Liu, 3D shape segmentation using soft density peak clustering and semi-supervised learning, Comput. Aided Des. 145 (2022) 103181.

[213] Ahmed Abdelreheem, Ivan Skorokhodov, Maks Ovsjanikov, Peter Wonka, SATR: Zero-shot semantic segmentation of 3D shapes, 2023, arXiv preprint arXiv:2304.04909.

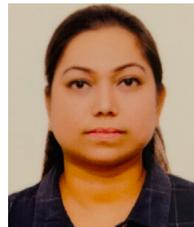

**Dr. Anju Rani** received an M.Tech from Thapar University, India and a Ph.D. from the Department of Electrical Engineering, Indian Institute of Technology Ropar, India. She is currently a post-doctorate in the Department of Energy at Aalborg University, Denmark. Her research interests include thermal non-invasive/non-destructive imaging technologies, condition monitoring, deep learning and bio-medical image processing.

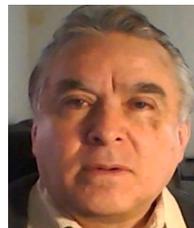

**Dr. Daniel Ortiz-Arroyo** is currently an associate professor in the Department of Energy at Aalborg University in Denmark. He earned a Ph.D. in Computer Science and Engineering at Oregon State University, USA. He has edited several books and is the author of more than 70 papers in international journals and conferences. His areas of research are in AI and Machine Learning, Robotics, Fuzzy Logic, NLP, security, computer architecture, and social network analysis.

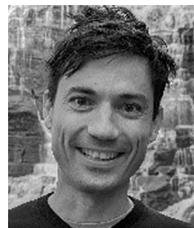

**Dr. Petar Durdevic** received a BSc in data and electronics and an MSc degree in control systems from Aalborg University, Denmark, in 2011 and 2013, respectively. He received his Ph.D. in Control Engineering from Aalborg University, Denmark in 2017. He is currently an associate professor at the Department of Energy Technology at Aalborg University, Denmark. His current research interests include nonlinear system identification and control, artificial intelligence, and condition monitoring with application domains in robotics and industrial processes.